%% file: main.tex
\documentclass[sigconf]{acmart}

\copyrightyear{2026}
\acmYear{2026}
\setcopyright{cc}
\setcctype{by}
\acmConference[WWW '26] {Proceedings of the ACM Web Conference 2026}{April 13--17, 2026}{Dubai, United Arab Emirates.}
\acmBooktitle{Proceedings of the ACM Web Conference 2026 (WWW '26), April 13--17, 2026, Dubai, United Arab Emirates}
\acmISBN{979-8-4007-2307-0/2026/04}
\acmDOI{10.1145/3774904.3792181}

\input{setup}

\title{Off-Policy Learning with Limited Supply}
\renewcommand{\shorttitle}{Off-Policy Learning with Limited Supply}

\author{Koichi Tanaka}
\authornote{Equal contribution.}
\affiliation{
  \institution{Keio University}
  \city{Tokyo}
  \country{Japan}
}
\email{kouichi_1207@keio.jp}

\author{Ren Kishimoto}
\authornotemark[1]
\affiliation{
  \institution{Institute of Science Tokyo}
  \city{Tokyo}
  \country{Japan}
}
\email{kishimoto.r.ab@m.titech.ac.jp}

\author{Bushun Kawagishi}
\affiliation{
  \institution{Meiji University}
  \city{Tokyo}
  \country{Japan}
}
\email{ee227051@meiji.ac.jp}

\author{Yusuke Narita}
\affiliation{
  \institution{Yale University}
  \city{New Haven, CT}
  \country{USA}
}
\email{yusuke.narita@yale.edu}

\author{Yasuo Yamamoto}
\affiliation{
  \institution{LY Corporation}
  \city{Tokyo}
  \country{Japan}
}
\email{yasyamam@lycorp.co.jp}

\author{Nobuyuki Shimizu}
\affiliation{
  \institution{LY Corporation}
  \city{Tokyo}
  \country{Japan}
}
\email{nobushim@lycorp.co.jp}

\author{Yuta Saito}
\affiliation{
  \institution{Hanjuku-kaso, Co., Ltd.}
  \city{Tokyo}
  \country{Japan}
}
\email{saito@hanjuku-kaso.com}

\renewcommand{\shortauthors}{Koichi Tanaka et al.}

\begin{document}

\input{draft_arxiv}

\bibliographystyle{ACM-Reference-Format}
\balance
\bibliography{ref}

\input{appendix}
\end{document}

%% file: setup.tex
\usepackage{amsthm}
\usepackage{amsmath}
\usepackage{bbm}
\usepackage{booktabs} 
\usepackage{multirow} 
\usepackage{mathtools}
\usepackage{graphicx}
\usepackage{subcaption}
\usepackage{algorithm}
\usepackage{algorithmic}
\usepackage{tcolorbox}
\usepackage{wrapfig}
\usepackage{cite}
\usepackage{cancel}
\theoremstyle{plain}

\newtheorem{theorem}{Theorem}[section]

\DeclareMathOperator*{\argmax}{arg\,max}

\newcommand{\calD}{\mathcal{D}}
\newcommand{\calX}{\mathcal{X}}

\newcommand{\calA}{\mathcal{A}}

\newcommand{\indicator}[1]{\mathbb{I}\{#1\}}

%% file: draft_arxiv.tex
\begin{abstract}
We study off-policy learning (OPL) in contextual bandits, which plays a key role in a wide range of real-world applications such as recommendation systems and online advertising. Typical OPL in contextual bandits assumes an unconstrained environment where a policy can select the same item infinitely. However, in many practical applications, including coupon allocation and e-commerce, limited supply constrains items through budget limits on distributed coupons or inventory restrictions on products. In these settings, greedily selecting the item with the highest expected reward for the current user may lead to early depletion of that item, making it unavailable for future users who could potentially generate higher expected rewards. As a result, OPL methods that are optimal in unconstrained settings may become suboptimal in limited supply settings. To address the issue, we provide a theoretical analysis showing that conventional greedy OPL approaches may fail to maximize the policy performance, and demonstrate that policies with superior performance must exist in limited supply settings. Based on this insight, we introduce a novel method called \emph{Off-Policy learning with Limited Supply} (OPLS). Rather than simply selecting the item with the highest expected reward, OPLS focuses on items with relatively higher expected rewards compared to the other users, enabling more efficient allocation of items with limited supply. Our empirical results on both synthetic and real-world datasets show that OPLS outperforms existing OPL methods in contextual bandit problems with limited supply.
\end{abstract}

\ccsdesc[500]{Information systems~Recommender systems}
\ccsdesc[300]{Information systems~Personalization}

\keywords{Off-Policy Learning, Contextual Bandits, Limited Supply.}

\maketitle

\begin{table*}[h]
\centering
\caption{Coupon allocation example: comparison between greedy and optimal methods. 
The left table shows $q(x,a)$ for each user–coupon pair. 
The right table illustrates allocations under different arrival orders, along with the policy  value $V(\pi)$.}
\begin{minipage}[c]{0.3\linewidth}
  \centering
  \scalebox{1.0}{
  \begin{tabular}{c|c|c|c}
    \toprule
     $q(x,a)$ & 30\%OFF & 50\%OFF & 70\%OFF \\
    \midrule
    $x_1$ & 80 & 250 & 200 \\
    $x_2$ & 100 & 280 & 120 \\
    $x_3$ & 60 & 100 & 70 \\
    \bottomrule
  \end{tabular}
  }
\end{minipage}
\hspace{0.5cm} 
\begin{minipage}[c]{0.6\linewidth}
  \centering
  \scalebox{1.0}{
    \begin{tabular}{c|cc}
    \toprule
    Arrival order & \hspace{0.8em}greedy & optimal  \\
    \midrule
    $(x_1,x_2,x_3)$ & 50\%\text{OFF}, 70\%\text{OFF}, 30\%\text{OFF} &  70\%\text{OFF}, 50\%\text{OFF}, 30\%\text{OFF} \\
    $(x_1,x_3,x_2)$ & 50\%\text{OFF}, 70\%\text{OFF}, 30\%\text{OFF} & 70\%\text{OFF},  30\%\text{OFF}, 50\%\text{OFF}\\
    $\cdots$ & $\cdots$  &  $\cdots$ \\
    $(x_3,x_2,x_1)$ & 50\%\text{OFF}, 70\%\text{OFF}, 30\%\text{OFF} & 30\%\text{OFF}, 50\%\text{OFF},  70\%\text{OFF} \\
    \midrule
    $V(\pi)$ & 420 & 540 \\
    \bottomrule
    \end{tabular}
  }
\end{minipage}
\label{table:coupon_example}
\end{table*}

\section{Introduction}

Decision-making problems play a crucial role in various application domains such as recommender systems~\citep{gilotte2018offline, saito2021counterfactual, felicioni2022off, saito2021evaluating, saito2021open}, online advertising~\citep{bottou2013counterfactual}, and search~\citep{li2015counterfactual}, where the use of data-driven algorithms has become increasingly important in recent years. These are often formulated as contextual bandit problems, where policies select an action (e.g., products, movies) based on the observed context (e.g., user feature) and receive a reward (e.g., click, conversion). The goal of the contextual bandit problem is to learn a policy that maximizes the expected reward. Off-policy learning (OPL)~\citep{swaminathan2015batch,metelli2021subgaussian, swaminathan2017off,saito2024long} enables policy learning solely from logged data, without deploying new policies online. This is particularly valuable because it allows for policy improvement without risking user experience or incurring the high cost of online experiments~\citep{kiyohara2024towards, saito2021counterfactual}. Typical OPL methods in contextual bandits assume an unconstrained environment where a policy can select each item infinitely. However, in many real-world applications, such as coupon allocation or e-commerce, popular or scarce items may run out over time or become unavailable because of limited supply \citep{zhong2015stock, christakopoulou2017recommendation}. Although conventional approaches that greedily aim to maximize the expected reward for each incoming user are optimal under typical OPL settings, they can become suboptimal with limited supply. This is because selecting an item that maximizes expected reward for a user at each time step may deplete its inventory, preventing it from being allocated to another user in the future who would generate a higher expected reward. 

To illustrate this issue, we present a simple coupon allocation example in Table~\ref{table:coupon_example}. There is exactly one coupon available for each discount type: “30\% OFF,” “50\% OFF,” and “70\% OFF.” We assume that three unique users ($x_1, x_2, x_3$) arrive sequentially, each visiting once. Let $q(x, a)$ denote the expected reward to the platform for each user–coupon pair $(x, a)$. The optimal allocation that maximizes the total expected reward assigns the 70\% OFF coupon to $x_1$ (reward 200), the 50\% OFF coupon to $x_2$ (reward 280), and the 30\% OFF coupon to $x_3$ (reward 60), yielding a total expected reward of 540 regardless of the arrival order. In contrast, conventional greedy methods allocate to each user the item that yields the highest expected reward among the remaining options. Since, for every user, the platform’s expected rewards are ranked in descending order as 50\% OFF, 70\% OFF, and 30\% OFF, the greedy method allocates coupons in this order as users arrive. As a result, the total reward heavily depends on the arrival order. When we assume that all possible arrival orders occur uniformly at random, the policy performance of the greedy method is 420, denoted in the table as \emph{greedy} (See Appendix~\ref{app:average-greedy} for detailed calculations.). This value is strictly lower than the optimal policy performance of 540, demonstrating that a greedy method, which is optimal in unconstrained settings, fails to achieve optimality under inventory constraints.

To formally consider such supply limitations, we formulate the problem of \textit{OPL with limited supply} for the first time in the relevant literature. To do so, we start by analyzing a simplified setting where each item is available in a single unit and all users share the same preference ordering over items. We theoretically prove that there always exists a policy that is comparable to or superior to the model-based greedy approach in terms of the policy performance under limited supply. Based on this analysis, we propose a novel method called \emph{Off-Policy learning with Limited Supply} (OPLS). Instead of selecting items greedily based on their expected rewards, OPLS selects items based on the \textit{relative reward gap}, calculated as a user's expected reward minus the average expected reward across all users. This approach prioritizes allocating items to users who yield relatively higher expected rewards compared to the other users. Importantly, OPLS requires no additional computational cost compared to existing model-based methods. 

In more realistic settings, items often differ substantially in their demand–supply conditions. For example, in e-commerce, popular or trending items often have limited supply relative to their demand, whereas long-tail products usually keep sufficient stock. Similarly, in coupon allocation, providers usually issue only a few high-discount coupons due to cost constraints, while they distribute low-discount coupons in large quantities. To handle even such heterogeneity, we further extend OPLS by applying different decision rules to different item groups:
\begin{itemize}
    \item \textbf{Limited supply items}, which will be sold out by the end of the horizon. 
    For these, we select the item with the highest \emph{relative} expected reward across users.
    \item \textbf{Abundant supply items}, which will remain in stock by the end of the horizon.
    For these, we select the item with the highest \emph{absolute} expected reward.
\end{itemize}
Finally, between the two item groups, OPLS chooses the one with the higher expected reward. This simple yet effective mechanism integrates the greedy strategy, which performs optimally when supply is abundant, and adapts to cases with limited supply, enabling OPLS to outperform the greedy method across a wide range of inventory conditions.
Through both synthetic and real-world experiments, we demonstrate that OPLS achieves high policy performance in contextual bandit problems with limited supply, whereas existing methods perform poorly under such constraints.

\section{Off-Policy Learning without Limited Supply}
This section formulates the standard problem setting for off-policy learning (OPL) in contextual bandits without limited supply \citep{saito2022off, saito2023off}. 
Let $x \in \mathcal{X} \subseteq \mathbb{R}^{d_x}$ represent a context vector (e.g., user feature). Given a context $x$, a possibly stochastic policy $\pi(a | x)$ selects an action $a \in \mathcal{A}$ (e.g., products, movies). Let $r \in \mathbb{R}_{\geq 0}$ denote a reward (e.g., a click or conversion), drawn from an unknown conditional distribution $p(r | x, a)$. In the OPL setting, we use a logged dataset $\mathcal{D} := \{(x_i, a_i, r_i)\}_{i=1}^n$ collected under the logging policy $\pi_0$ and the generative process produces each tuple according to the distribution $(x, a, r) \sim p(x)\pi_0(a | x)p(r | x, a)$. The expected reward under a policy $\pi$, known as the \textit{policy value}, is defined as
\begin{align*} 
    V(\pi) := \mathbb{E}_{p(x)\pi(a|x)p(r|x, a)}[r] = \mathbb{E}_{p(x)\pi(a|x)}[q(x, a)], 
\end{align*} 
where $q(x, a) := \mathbb{E}[r | x, a]$ is the expected reward function, also known as the \textit{q-function}. The objective in OPL is to learn a policy $\pi$ that maximizes the policy value using only the logged dataset $\pi = \argmax V(\pi).$

One commonly used method in OPL is the greedy model-based approach, which first estimates the $q$-function $q(x, a)$ using standard supervised learning techniques. From the estimated function $\hat{q}(x, a)$, we typically obtain a policy by taking the action that maximizes the estimated reward
\begin{align}
    \pi(a | x) =
    \begin{cases}
        1 & \text{if } a = \argmax_{a' \in \mathcal{A}} \hat{q}(x, a') \\
        0 & \text{otherwise} 
    \end{cases}
    \label{eq:reg}.
\end{align}
When we know the true $q$-function, this approach can always select the optimal action $a$ for a given context $x$, thereby yielding the optimal policy. In practice, however, it may suffer from bias due to estimation errors in the $q$-function, yet it remains one of the simplest and most computationally efficient approaches \citep{saito2021counterfactual, saito2024potec}.

So far, we have discussed the standard formulation of OPL. This formulation assumes no constraints on item availability, meaning that a policy can select each item infinitely if needed.
However, in many real-world scenarios, such as coupon allocation,  e-commerce, popular or scarce items may run out over time or become unavailable because of limited supply. Therefore, we next consider the off-policy learning with limited supply.

\section{Related Work}
This section introduces key related work.
\paragraph{\textbf{Off-Policy Evaluation and Learning}}
Off-Policy Evaluation (OPE) and Off-Policy Learning (OPL)~\citep{dudik2014doubly, wang2017optimal, liu2018breaking, farajtabar2018more, su2019cab, su2020doubly, kallus2020optimal, metelli2021subgaussian, saito2022off, saito2023off} aim to estimate or optimize a policy using only logged data. 
A variety of methods and estimators have been proposed in OPE to control the bias–variance tradeoff when estimating the policy value ~\citep{su2020doubly, su2019cab, swaminathan2015counterfactual, swaminathan2015self, dudik2014doubly, farajtabar2018more, kallus2020optimal, kiyohara2022doubly, kiyohara2023off, metelli2021subgaussian, saito2021counterfactual, wang2017optimal, kiyohara2024off}.
These techniques achieve accurate policy value estimation in standard settings without limited supply~\citep{uehara2022review,saito2021evaluating,saito2021open}, and have been extended in OPL to estimate policy gradients and improve learning from logged data \citep{saito2023off}. 
While many of these methods focus on standard contextual bandit settings, real-world applications often involve a limited supply. Our work is the first to formulate OPL with inventory constraints in contextual bandits, providing a principled framework that existing methods cannot directly handle.

\paragraph{\textbf{Recommendation with Constraints}}

Several prior studies have investigated recommendations under budget or inventory constraints. Bandits with Knapsacks (BwK) framework~\citep{badanidiyuru2018bandits, agrawal2016linear} models multi-armed bandit problems with budget constraints. 
Since BwK does not utilize user features, it cannot provide personalized recommendations. Resourceful Contextual Bandits~\citep{badanidiyuru2014resourceful} extend this setting by incorporating contextual information, allowing for more personalized decisions. Nevertheless, due to computational limitations, it is known to be challenging to apply these methods directly for personalized recommendation~\citep{wu2015algorithms, zhao2024constrained}. Importantly, these approaches primarily consider global budget constraints, where limitations apply across all users and items, rather than per-item constraints. In contrast, our setting focuses on inventory constraints at the item level. 
Relatively few existing studies address inventory-constrained recommendations~\citep{christakopoulou2017recommendation, zhong2015stock}. ~\citet{christakopoulou2017recommendation} proposes a matrix factorization-based approach that ensures item recommendations do not exceed item capacity, but their setting does not involve sequential user arrival as in the bandit framework. ~\citet{zhong2015stock} proposes a method for learning recommendations that ensure item sales remain within inventory limits. While this method considers per-item inventory constraints, it does not incorporate user features. Our work, in contrast, explicitly considers user features to determine which item to recommend to which user, given the remaining inventory.
Finally, while most of the aforementioned studies focus on online learning, our work considers the offline setting. To the best of our knowledge, this is the first study to introduce a limited supply into OPL for contextual bandits.

\section{Off-Policy Learning with Limited Supply}
While the previous section focused on OPL in conventional contextual bandit settings, in this section, we introduce a novel formulation of the OPL problem with limited supply. In this setting, each item has an inventory that evolves over time, and the set of available actions changes depending on the remaining inventory. Consequently, we must redefine both the logged data and the policy value to properly capture the dynamics of limited supply.
 
We consider a recommendation horizon of $T$ rounds, during which the policy makes recommendations sequentially. In contrast to standard contextual bandits, we introduce two new variables, $s$ and $c$, in order to represent the transitions of inventory. Let $s$ represent the inventory state, where $s_t \in \mathbb{R}^{|\mathcal{A}|}$ is a vector whose each element $s_t^a$ denotes the inventory count of an item $a$ at time $t$. Let $c \in \{0, 1\}$ be a binary inventory signal that indicates whether the system consumes the inventory. For example, in e-commerce, $c = 1$ when a user purchases an item and its inventory decreases. In coupon allocation, $c = 1$ if the platform distributes a coupon. At time $t$, the platform deterministically updates the inventory from the previous state and the binary inventory signal through a known transition function $p(s' \mid s, a, c)$. Specifically, the platform updates the inventory of item $a$ at time $t$ as follows

\begin{align}
    s_t^a = s_1^a - \sum_{t' < t} c_{t'} \cdot \mathbb{I}\{a_{t'} = a\}.
\end{align}
Furthermore, we define the policy as $\pi(a | x, s)$, which selects actions conditioned on both context $x$ and the current inventory $s$. Given these definitions, we formulate the logged dataset as 
\begin{align*}
\mathcal{D} := \left\{\left\{(x_{i,t}, a_{i,t}, c_{i,t}, r_{i,t}, s_{i,t})\right\}_{t=1}^T\right\}_{i=1}^n,
\end{align*}
which is collected under a logging policy $\pi_0$, and thus the generative process samples each tuple 
\begin{align*}
(x, a, c, r, s) \sim p(x)\, \pi_0(a | x, s)\, p(c | x, a)\, p(r | x, a)\, p(s' |s, a, c).
\end{align*}

We consider a reward model where we obtain rewards only when the item is actually consumed ($c = 1$), such as when a purchase generates revenue or distributing a coupon produces its effect. Hence, we define the policy value as the expected value of the product $c \cdot r$ across all time steps:
\begin{align*}
    V(\pi) &:= \mathbb{E}_{\substack{p(s_1)\,\prod_{t=1}^T p(x_t)\pi(a_t|x_t,s_t)p(c_t|x_t,a_t)\\ \quad \qquad p(r_t|x_t,a_t)p(s_{t+1}|s_t, a_t, c_t)}}
   \left[\sum_{t=1}^T c_t \cdot r_t\right].
\end{align*}
Since $c$ is binary and the transition $p(s' |s, a, c)$ is deterministic, we then derive it as
\begin{align}
    V(\pi) &= \sum_{t=1}^T \mathbb{E}_{p(s_1) p(x_t) \pi(a_t | x_t, s_t) p(c_t | x_t, a_t) p(r_t | x_t, a_t)}\left[ c_t \cdot r_t \right] \notag\\
    &= \sum_{t=1}^T \mathbb{E}_{p(s_1) p(x_t) \pi(a_t | x_t, s_t)}\left[ q_c(x_t, a_t) \cdot q_r(x_t, a_t) \right] \notag\\
    &= \sum_{t=1}^T \mathbb{E}_{p(s_1) p(x_t) \pi(a_t | x_t, s_t)}\left[ q(x_t, a_t) \right],
    \label{eq:V_pi_modified}
\end{align}
where $q_c(x, a) := \mathbb{E}_{p(c | x, a)}[c]$, $q_r(x, a) := \mathbb{E}_{p(r | x, a)}[r]$, and $q(x, a) := \mathbb{E}_{p(c | x, a)p(r | x, a)}[c \cdot r]$. See Appendix \ref{derivation of policy value} for detailed derivations.

As in standard OPL, the goal of \textit{OPL with limited supply} (our setup) is to learn a policy $\pi$ that maximizes the policy value $V(\pi)$. However, the critical difference lies in the dependence of $\pi$ on the inventory state $s_t$, as seen in $\pi(a | x, s_t)$. In the next section, we describe algorithms for solving this problem, beginning with adaptations of existing OPL methods to the limited supply setting.

\subsection{Conventional Greedy Approach}

Since no existing methods address the limited supply setting, we first extend the standard model-based approach to the settings with limited supply. Let $\mathcal{A}_{s_t} := \{a \in \mathcal{A} \mid s_t^a > 0\}$ denote the set of actions with remaining inventory at time $t$. Then, Eq.~\eqref{eq:reg} can be extended to the limited supply setting as
\begin{align}
    \pi_{\text{greedy}}(a_t | x_t, s_t) =
    \begin{cases}
        1 & \text{if } a_t = \argmax_{a \in \mathcal{A}_{s_t}} \hat{q}(x_t, a) \\
        0 & \text{otherwise}
    \end{cases}
    \label{eq:reg_supply}.
\end{align}

This policy greedily selects the item with the highest predicted reward $\hat{q}(x, a)$ among those with available inventory $\mathcal{A}_{s_t}$. In typical contextual bandit settings, this greedy approach is optimal when $\hat{q}(x, a)$ accurately estimates the true expected reward function. However, with limited supply, such greedy selection can be suboptimal. If the policy allocates the item that maximizes the expected reward to a user at each time step, the item may become unavailable to another user in the future who could generate a higher expected reward. As discussed in the introduction section with the coupon allocation example in Table~\ref{table:coupon_example}, the conventional greedy method performs worse than the optimal policy. This motivates the development of novel \textit{inventory-aware} OPL approaches as our main contribution.

\begin{table*}[h]
\centering
\caption{Coupon allocation example: OPLS allocation. 
The left table shows the relative reward gap $q(x,a) - \mathbb{E}[q(x,a)]$ for each user–coupon pair. The right table illustrates allocations under different arrival orders, along with the policy value $V(\pi)$.}

\begin{minipage}[c]{0.5\linewidth}
  \centering
  \scalebox{1.0}{
  \begin{tabular}{c|c|c|c}
    \toprule
     $q(x,a) - \mathbb{E}[q(x, a)]$ & 30\%OFF & 50\%OFF & 70\%OFF \\
    \midrule
    $x_1$ & 0 (80 - 80)& 40 (250 -210) & \textbf{70} (200 -130)\\
    $x_2$ & 20 (100 - 80) & \textbf{70} (280 -210) & -10 (120 -130)\\
    $x_3$ & \textbf{-20} (60 - 80) & -110 (100 -210) & -60 (70 -130)\\
    \bottomrule
  \end{tabular}
  }
\end{minipage}
\hspace{0.5cm} 
\begin{minipage}[c]{0.4\linewidth}
  \centering
  \scalebox{1.0}{
    \begin{tabular}{c|c}
    \toprule
    Arrival order & OPLS  \\
    \midrule
    $(x_1,x_2,x_3)$  &  70\%\text{OFF}, 50\%\text{OFF}, 30\%\text{OFF} \\
    $(x_1,x_3,x_2)$  & 70\%\text{OFF},  30\%\text{OFF}, 50\%\text{OFF}\\
    $\cdots$   &  $\cdots$ \\
    $(x_3,x_2,x_1)$ & 30\%\text{OFF}, 50\%\text{OFF},  70\%\text{OFF} \\
    \midrule
    $V(\pi)$  & 540 \\
    \bottomrule
    \end{tabular}
  }
\end{minipage}
\label{table:coupon_example_opls}
\end{table*}

\subsection{The Proposed Method}

To design an effective method to deal with the problem of OPL with limited supply, we begin by comparing the optimal policy with the greedy model-based policy. We first present a theoretical result under a simplified setting.

\begin{tcolorbox}[colback=gray!10]
\begin{theorem}
Let there be $J$ users $x_1, \ldots, x_J$ and $K$ actions $a_1, \ldots, a_K$, where each action has exactly one unit of inventory. Assume the reward function satisfies
\begin{align*}   
    q(x_j, a_k) \geq q(x_j, a_{k+1})
\end{align*}
for all $j$ and $k$. Then, for any user $x_j$ and item $a_k$, the following inequality holds:
\begin{align}
    &V(\pi^*) - V(\pi_{\text{greedy}}) \notag\\
    &\geq p(x_j) \Big\{ 
        \left(q(x_j, a_k) - \mathbb{E}_{p(x)}[q(x, a_k)]\right) \notag\\
        & \qquad \qquad - \left(q(x_j, a_1) - \mathbb{E}_{p(x)}[q(x, a_1)]\right) 
    \Big\}
    \label{eq: difference optimal}
\end{align}
where $V(\pi^*)$ denotes the value of the optimal policy and $V(\pi_{\text{greedy}})$ is the value of the greedy model-based policy.
\label{theorem:difference optimal}
\end{theorem}
\end{tcolorbox}
This theorem considers a simplified setting where each item has exactly one unit in stock and all users share the same preference order over items. It provides a lower bound on the difference in policy values between the optimal policy and the greedy model-based policy. To prove this result, we consider the difference in policy values between the greedy model-based policy $\pi_{\text{greedy}}$ and a modified policy $\pi_{j,k}$ that selects item $a_k$ for user $x_j$ at time $t=1$ and follows the greedy model-based policy thereafter. We derive the policy value of the greedy model-based policy as
\begin{align}
    V(\pi_{\text{greedy}}) = \sum_{k=1}^K \mathbb{E}_{p(x)}[q(x, a_k)]. 
    \label{eq:V_pi_greedy}
\end{align}
We also derive the policy value of $\pi_{j,k}$ as
\begin{align}
    V(\pi_{j,k}) &= p(x_j) \left\{ q(x_j,a_k)-\mathbb{E}_{p(x)}[q(x,a_k)]  \right.\notag \\
    & \left. \quad - (q(x_j,a_1) - \mathbb{E}_{p(x)}[q(x,a_1)]) \right\} + V(\pi_{\text{greedy}}),
    \label{eq:V_pi_jk}
\end{align}
which yields the result in Theorem~\ref{theorem:difference optimal}, and we provide the detailed proof in Appendix~\ref{derivation of greedy} and \ref{derivation of pi_jk}. Setting $k=1$ in Eq.\eqref{eq: difference optimal} yields a lower bound of zero, implying that there always exists a policy that outperforms or performs equally well as the greedy model-based policy.

The key insight for obtaining a policy that performs better than the greedy policy comes from maximizing the lower bound in Eq.\eqref{eq: difference optimal}. 
To maximize this bound, it is desirable to choose an item $a_k$ that maximizes 
$q(x_j, a_k) - \mathbb{E}_{p(x)}[q(x, a_k)]$. 
In other words, it is important to allocate items to users who achieve relatively higher expected rewards compared to others, which is intuitively reasonable as well. To realize this idea, we propose a new method called \emph{Off-Policy Learning with Limited Supply} (OPLS), 
which selects items based on the \textit{relative reward gap}, defined as
\[
    \hat{q}(x, a) - \frac{1}{n} \sum_{i=1}^n \hat{q}(x_i, a),
\]
that is, a user’s expected reward minus the average expected reward across all users.
Formally, we define OPLS as
\begin{align*}
    \pi_{\text{OPLS}}(a_t | x_t, s_t) =  
    \left\{
    \begin{aligned}
        1  \quad & \text{if } a_t = \underset{a \in \mathcal{A}_{s_t}}{\operatorname{argmax}} \left\{ \hat{q}(x_t, a) - \frac{1}{n} \sum_{i=1}^n \hat{q}(x_i, a) \right\} \\
        0 \quad & \quad \text{otherwise}
    \end{aligned}
    \right. .
\end{align*}
This approach enables more effective allocation than the greedy method by considering which users receive the limited items. Table~\ref{table:coupon_example_opls} illustrates the case where we apply OPLS to the simple coupon allocation example. We compute the average expected reward for each coupon: 80 for the 30\% OFF coupon, 210 for the 50\% OFF coupon, and 130 for the 70\% OFF coupon. 
OPLS selects the coupon that maximizes the difference between a user’s expected reward and the average expected reward of that coupon. In this example, OPLS can make the same recommendation as the optimal allocation, regardless of the order in which users arrive. Additionally, it requires no additional computation beyond estimating $\hat{q}$, which can be learned using the same model as the conventional greedy approach.

\paragraph{Extension to Mixed Supply Conditions}
As we have discussed so far, we proposed OPLS as a method for handling limited supply conditions. 
In more realistic scenarios, however, items can differ significantly in their demand–supply balance. 
For instance, in e-commerce, popular or trending items often have limited supply relative to demand, whereas long-tail products typically remain well-stocked. We can here extend OPLS to even effectively handle the challenging case where items differ in their demand–supply conditions. Let $\mathcal{A}_{\text{sold}}$ denote the set of items that will be sold out, and $\mathcal{A}_{\text{unsold}}$ denote the set of items that will remain in stock at time $T$, where
\begin{align*}
\mathcal{A}_{\text{sold}} = \{ a \in \mathcal{A} \mid s_T^a \leq 0 \}, \quad
\mathcal{A}_{\text{unsold}} = \{ a \in \mathcal{A} \mid s_T^a > 0 \}.
\end{align*}

For each group, we apply different decision rules.
\begin{itemize}
    \item \textbf{Limited supply items} ($a \in \mathcal{A}_{\text{sold}}$):  
    choose the item with the highest \emph{relative} expected reward across users,
    \[
        a_{\text{sold}, t} = \argmax_{a \in \mathcal{A}_{\text{sold}, t}}
        \left\{ \hat{q}(x_t, a) - \frac{1}{n}\sum_{i=1}^n \hat{q}(x_i, a) \right\}.
    \]
    \item \textbf{Abundant supply items} ($a \in \mathcal{A}_{\text{unsold}}$):  
    choose the item with the highest \emph{absolute} expected reward,
    \[
        a_{\text{unsold}, t} = \argmax_{a \in \mathcal{A}_{\text{unsold}, t}} \hat{q}(x_t, a).
    \]
\end{itemize}
Here, $\mathcal{A}_{\text{sold}, t} = \{ a \in \mathcal{A}_{\text{sold}} \mid s_t^a > 0 \}$ and $\mathcal{A}_{\text{unsold}, t} = \{ a \in \mathcal{A}_{\text{unsold}} \mid s_t^a > 0 \}$  denote the subsets of items with positive inventory at time $t$. We select the final action from these two candidates
\begin{align}
    \pi_{\text{OPLS}}(a_t | x_t, s_t) = \argmax_{a \in \{a_{\text{sold}, t}, a_{\text{unsold}, t}\}} \hat{q}(x_t, a).
\end{align}
This extension allows OPLS to greedily select the highest-reward items when they will remain in stock, while applying the relative-reward strategy when supply is scarce. OPLS thus integrates the conventional greedy method, achieving the same performance as greedy when items have sufficient inventory. This ensures that OPLS consistently matches or outperforms the greedy method under a wide range of inventory conditions.

To implement this procedure, we must estimate which items will be sold out and obtain $\mathcal{A}_{\text{sold}}$. At time $t$, we can express the probability that the policy consumes item $a$ as
\begin{align*}
    \pi^t(a) = \mathbb{E}_{p(x)}[\pi(a_t | x, s_t) \cdot q_c(x, a_t)],
\end{align*}
and the remaining inventory of item $a$ at time $T$ as
\begin{align*}
    s_T^a = s_0^a - \sum_{t=1}^T \pi^t(a).
\end{align*}
We can thus estimate the set of sold-out items as
$\mathcal{A}_{\text{sold}} = \{ a \in \mathcal{A}| s_T^a \leq 0 \}$.
However, since $\pi$ depends on $\mathcal{A}_{s_t}$ 
and evolves over time, estimating this set becomes computationally intractable when the number of items or the horizon $T$ is large.\footnote{Here, we use $\pi(a_t | x, s_t)$ from the base OPLS before the extension.} To overcome this difficulty, we also adopt a naive approximation in which we assume that the policy selects all items uniformly at each time step
\begin{align*}
    s_T^a = s_0^a - T \cdot \frac{1}{|\mathcal{A}|} \sum_{i=1}^n q_c(x_i, a).
\end{align*}
We empirically show that OPLS, even when using this naive estimation, consistently outperforms existing methods in scenarios where items have varying levels of supply in Appendix \ref{additional_result}.

\section{Synthetic Experiments}
This section empirically evaluates OPLS's performance using synthetic data and identifies the situations where OPLS is particularly more effective. 

\begin{figure*}[t]
  \centering
  \vspace{-18mm}
  \begin{minipage}[b]{0.55\linewidth}
    \centering
    \includegraphics[width=0.6\linewidth]{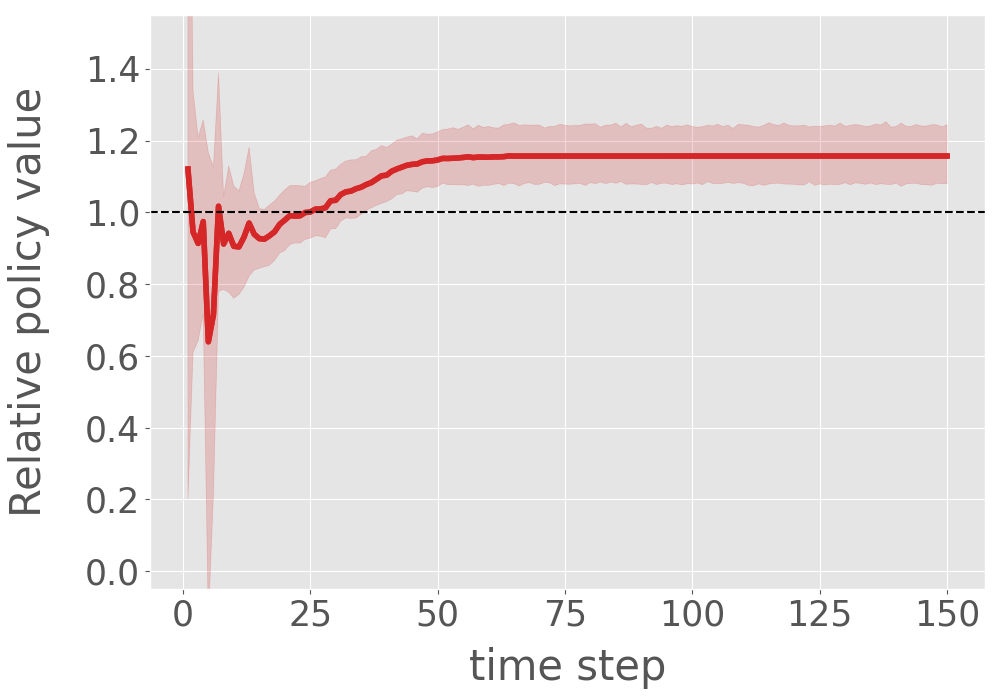}
    \caption{Relative policy value ($V(\pi_{\text{OPLS}})/V(\pi_{\text{greedy}})$) at each time step}
    \label{fig:small-scall_timestep}
  \end{minipage}
  \begin{minipage}[b]{0.4\linewidth}
  \vspace{15mm}
    \centering
    \makeatletter
    \def\@captype{table}
    \makeatother
    \caption{The expected reward function in the experiment of Figure~\ref{fig:small-scall_timestep}}
    \label{table:small-scale_q(x,a)}
    \vspace{3mm}
    \begin{tabular}{c|c|c|c|c|c}
    \toprule
     $q(x,a)$ & $a_1$ & $a_2$ & $a_3$ & $a_4$ & $a_5$ \\
    \midrule
    $x_1$ & 0.799 & 1.011 & 1.047 & 2.521 & 3.046 \\
    $x_2$ & 0.329 & 0.494 & 1.683 & 2.092 & 2.589 \\
    $x_3$ & 1.287 & 1.718 & 1.984 & 2.932 & 3.369 \\
    \bottomrule
  \end{tabular}
  \vspace{20mm}
  \end{minipage}
\end{figure*}

\begin{figure*}[t]
  \begin{subfigure}[b]{0.33\linewidth}
    \vspace{-3mm}
    \centering
    \includegraphics[width=0.85\linewidth]{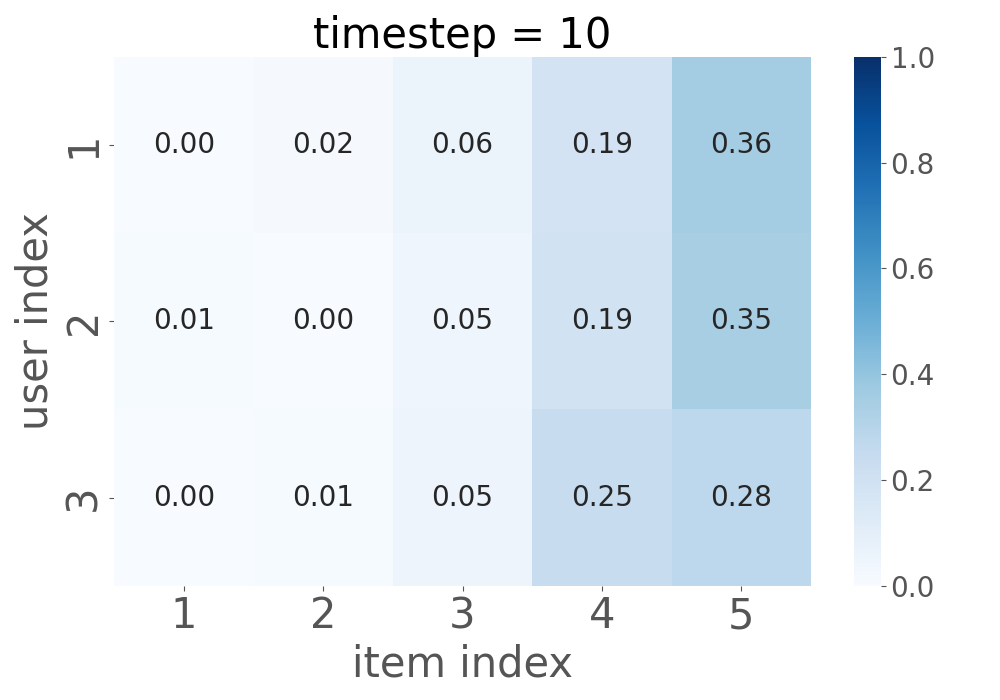}
    \vspace{-3mm}
  \end{subfigure}
  \begin{subfigure}[b]{0.33\linewidth}
    \centering
    \includegraphics[width=0.85\linewidth]{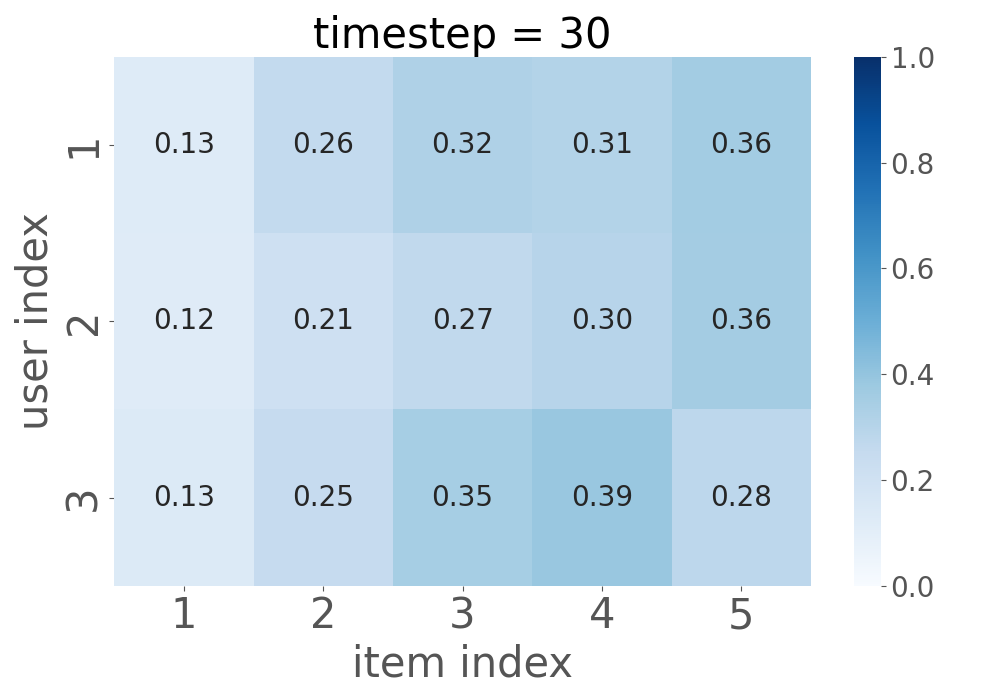}
    \vspace{-3mm}
  \end{subfigure}
  \begin{subfigure}[b]{0.33\linewidth}
    \centering
    \includegraphics[width=0.85\linewidth]{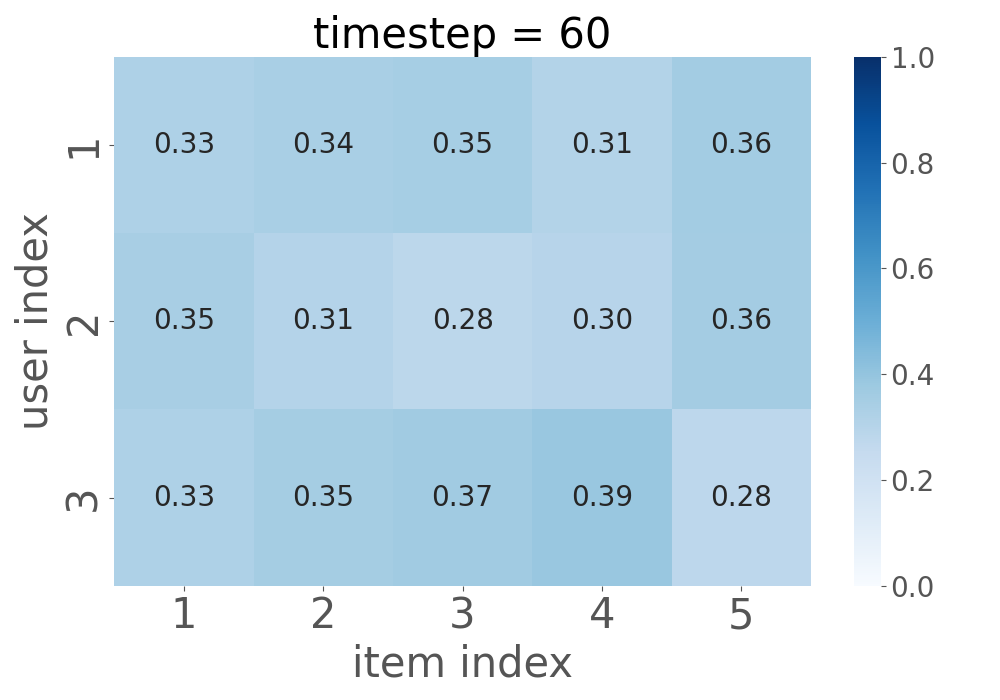}
    \vspace{-3mm}
  \end{subfigure}
  \caption{The behavior of the conventional greedy method at $t = 10, 30, 60$}
  \label{fig:behavior_argmax}
\end{figure*}

\begin{figure*}[t]
  \begin{subfigure}[b]{0.33\linewidth}
    \vspace{3mm}
    \centering
    \includegraphics[width=0.85\linewidth]{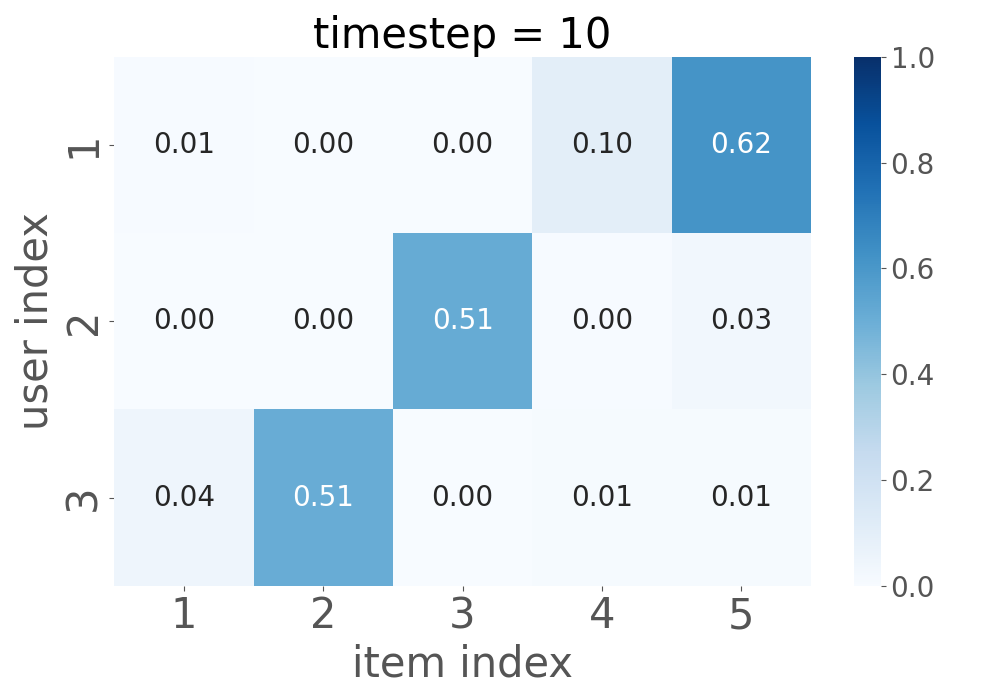}
    \vspace{-3mm}
  \end{subfigure}
  \begin{subfigure}[b]{0.33\linewidth}
    \centering
    \includegraphics[width=0.85\linewidth]{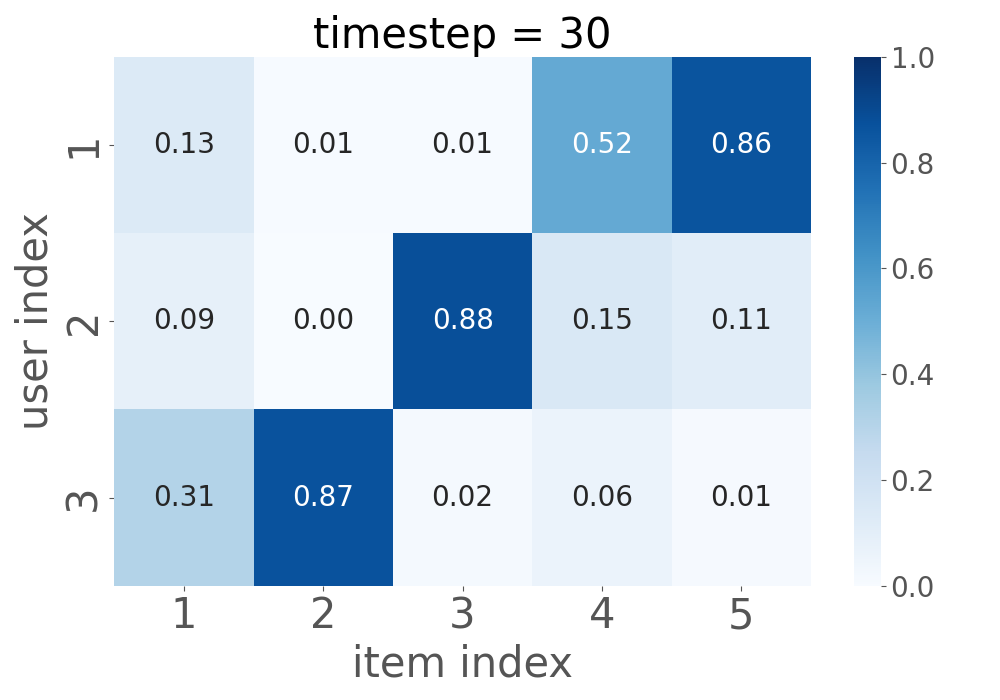}
    \vspace{-3mm}
  \end{subfigure}
  \begin{subfigure}[b]{0.33\linewidth}
   \centering
    \includegraphics[width=0.85\linewidth]{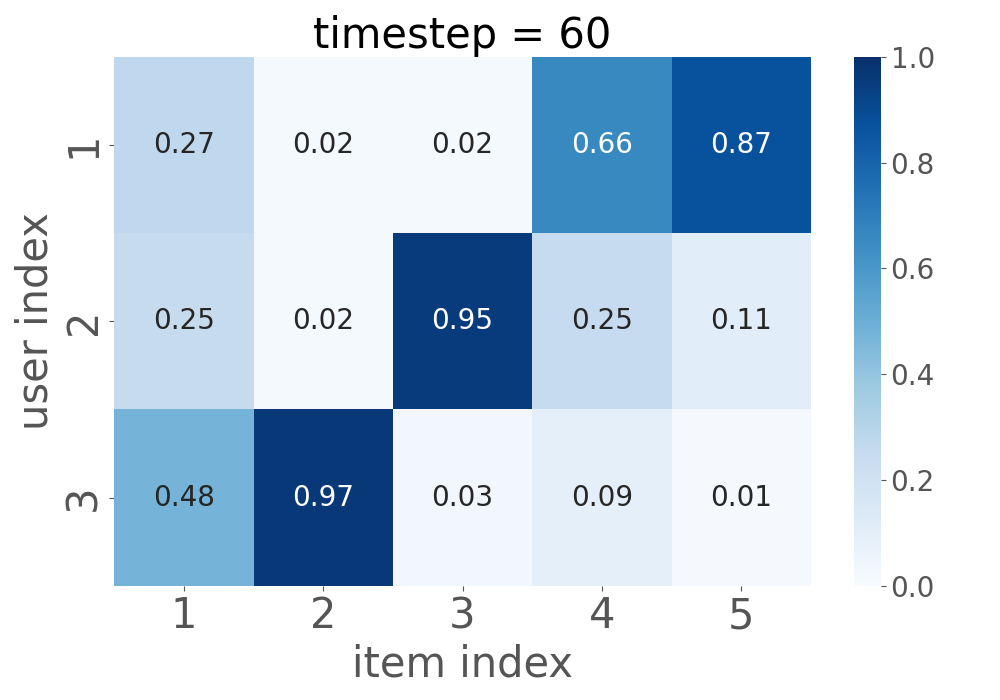}
    \vspace{-3mm}
  \end{subfigure}
  \caption{The behavior of OPLS at $t = 10, 30, 60$}
  \label{fig:behavior_OPLS}
\end{figure*}

\subsection{Setup}
To generate synthetic data, we first define 200 users characterized by 10-dimensional context vectors ($x$) sampled from the standard normal distribution. Then, we define a click probability $q_c(x,a)$ and an action value $q_r(x,a)$ as follows.
\begin{align}
    q_c(x,a) = \lambda \cdot f_c(x,a) + (1-\lambda) \cdot g_c(x,a) , \label{eq:p_c(x,a)}\\
    q_r(x,a) = \lambda \cdot f_r(x,a) + (1-\lambda) \cdot g_r(x,a), \label{eq:q_r(x,a)}
\end{align}
where $g(x,a)$ is a reward function that satisfies $g(x,a_k) \ge g(x,a_{k+1})$ for all users. We can control the popularity of actions among users by $\lambda$. If we set $\lambda = 0$, all users prefer the actions in the same order. Based on the above click probability, we sample a binary click signal $c$ from a binomial distribution whose mean is $q_c(x,a)$. We then sample a reward $r$ from a normal distribution whose mean is $q_r(x,a)$ and standard deviation $\sigma$ is 3.0.

We then synthesize the logging policy as follows.
\begin{align}
    \pi_0(a_t|x_t,s_t) = \frac{\exp{(\beta \cdot q(x_t,a_t))}}{\sum_{a' \in \calA_{s_t}} \exp{(\beta \cdot q(x_t,a'))}},
    \label{eq:pi_0}
\end{align}
where $\beta$ is an experimental parameter to control the stochasticity and optimality of the logging policy. We use $\beta = -1.0$ in the synthetic experiments.

To summarize, we first observe a user represented by $x$ and define the click probability and action value as in Eq.~\eqref{eq:p_c(x,a)} and Eq.~\eqref{eq:q_r(x,a)}. We then sample a discrete action $a$ from $\pi_0$ in Eq.~\eqref{eq:pi_0}. The click signal $c$ is sampled from a binomial distribution whose mean is $q_c(x,a)$ in Eq.~\eqref{eq:p_c(x,a)}. The reward $r$ is sampled from a normal distribution whose mean is $q_r(x,a)$ and standard deviation $\sigma$ is 3.0. If $c=1$, the supply of the recommended action reduces by one. Iterating this procedure $T$ times generates the logged data $\calD = \{ (x_t,a_t,c_t, r_t, s_t) \}_{t=1}^T$.

\paragraph{Initial Supply}
To identify the situation where OPLS provides more improvements, we define the initial supply $s_1$ by three different ways as follows. 

\begin{align*}
    s_1 = \left\{ 
  \begin{alignedat}{2}      
    &\quad s_{\text{max}} \cdot \frac{\mathbb{E}_{p(x)}[q(x,a)]}{\underset{a \in \mathcal{A}} {\operatorname{max}} \,\, \mathbb{E}_{p(x)}[q(x,a)]} \quad &\text{(proportional)} \\
    &\quad s_{\text{max}} \cdot \frac{\underset{a \in \mathcal{A}} {\operatorname{min}} \,\, \mathbb{E}_{p(x)}[q(x,a)]}{\sqrt{\mathbb{E}_{p(x)}[q(x,a)]}} \quad &\text{(inverse proportional)}\\
    &\quad \text{random.uniform(1, $s_{\text{max}}$)} \quad &\text{(random)}
  \end{alignedat} 
  \right.,
\end{align*}
where $s_{\text{max}}$ is the maximum value of supply. First, "proportional supply" simulates the situation where item supply increases proportionally with user demand. The more users want an item, the more supply of the item we have. Second, "inverse proportional supply" simulates the situation where item supply is inversely proportional to user demand. Items with low supply are typically in high demand. Finally, "random supply" defines initial supply at random.

\begin{figure*}[t]
    \centering
    \includegraphics[scale=0.30]{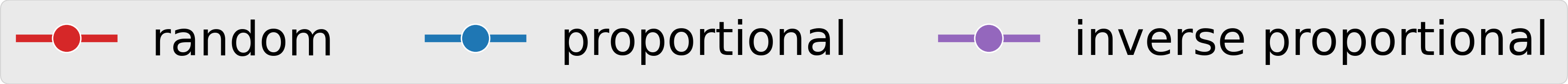} \\
    \Description{}
    \vspace{-3mm}
  \begin{subfigure}[b]{0.33\linewidth}
    \vspace{3mm}
    \centering
    \includegraphics[width=0.9\linewidth]{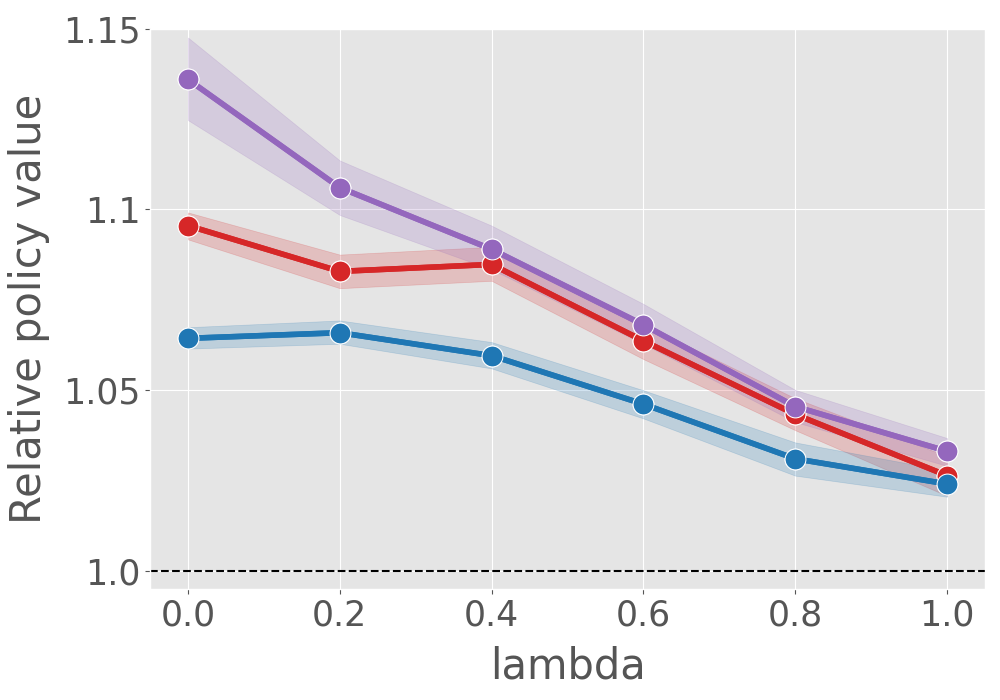}
    \vspace{-3mm}
    \caption{Relative policy values varying the action popularity ($\lambda$)}
    \label{fig:lambda}
  \end{subfigure}
  \begin{subfigure}[b]{0.33\linewidth}
    \centering
    \includegraphics[width=0.9\linewidth]{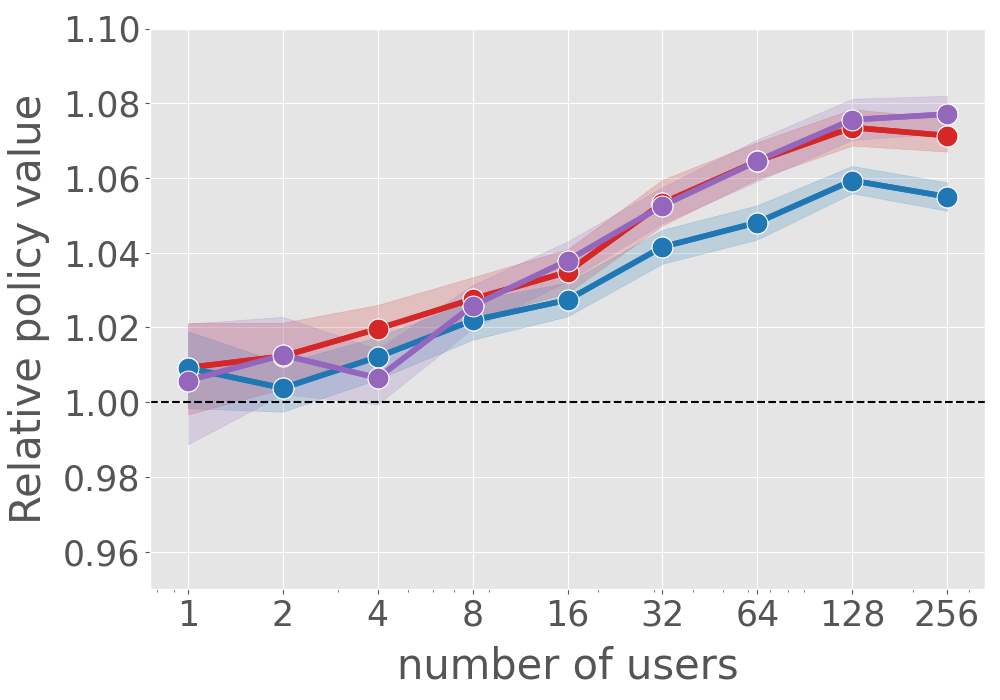}
    \vspace{-3mm}
    \caption{Relative policy values varying the number of users}
    \label{fig:user_action_ratio}
  \end{subfigure}
  \begin{subfigure}[b]{0.33\linewidth}
    \centering
    \includegraphics[width=0.9\linewidth]{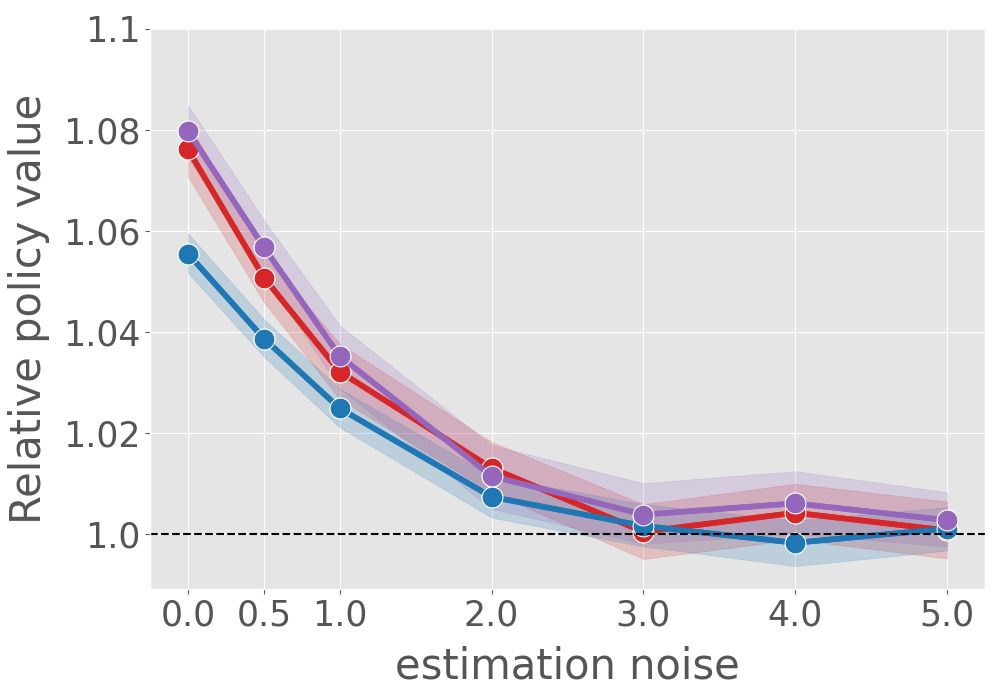}
    \vspace{-3mm}
    \caption{Relative policy values varying estimation noise}
    \label{fig:estimation_noise}
  \end{subfigure}
  \caption{Comparisons of relative policy values with varying (a)action popularity, (b)the number of users, and (c)estimation noise. The period $T$ is sufficiently large, and all items are sold out.}
  \label{fig:T=inf}
\end{figure*}

\begin{figure}
  \centering
  \vspace{-1mm}
  \includegraphics[scale=0.27]{image/legend.png} \\
  \vspace{1mm}
  \centering
  \includegraphics[width=0.7\linewidth]{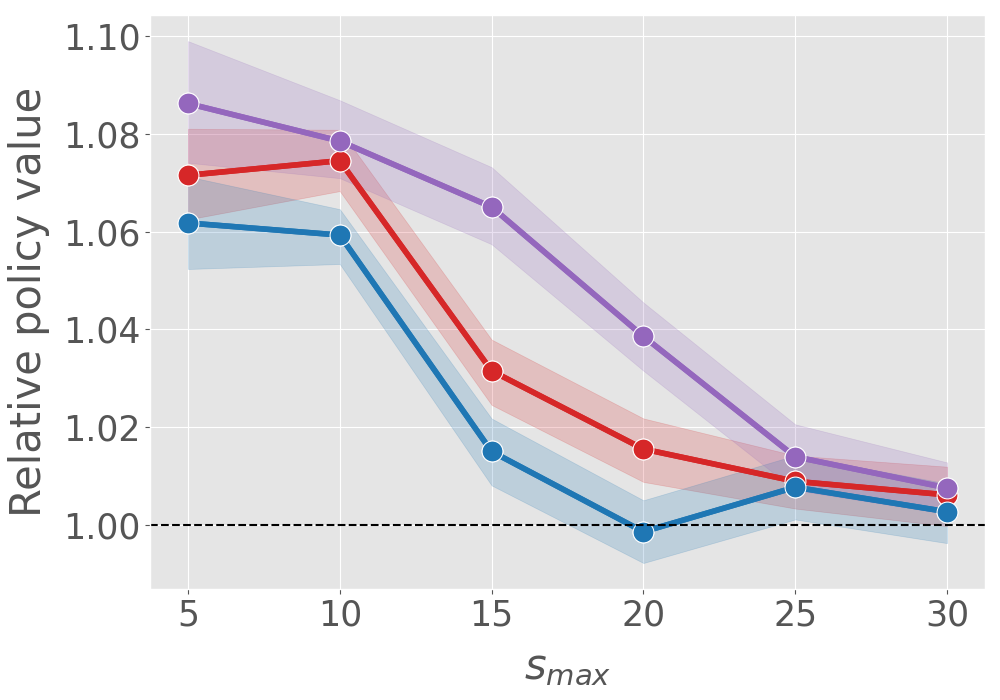} \vspace{-3mm}
  \Description{}
  \caption{Relative policy values varying max supply ($s_{max}$)}
  \vspace{-5mm}
  \label{fig:s_max}
\end{figure}

\begin{figure*}[t]
    \centering
    \includegraphics[scale=0.30]{image/legend.png} \\
    \vspace{-5mm}
    \Description{}
  \begin{subfigure}[t]{0.33\linewidth}
    \vspace{3mm}
    \centering
    \includegraphics[width=0.9\linewidth]{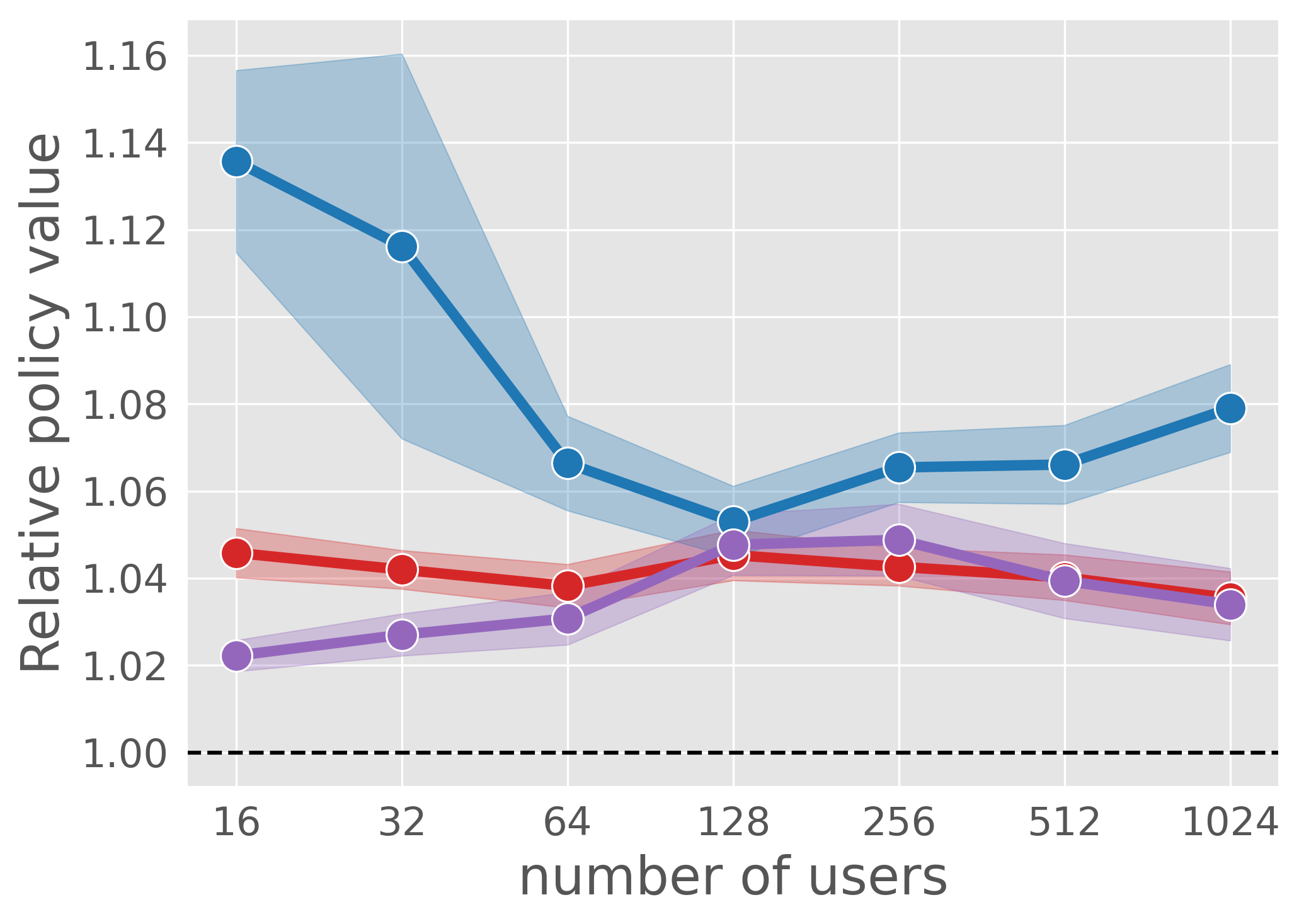}
    \vspace{-2mm}
    \caption{Relative policy values varying the number of users, using the estimated expected reward}
    \label{fig:n_users_estimate_KuaiRec}
  \end{subfigure}
  \hspace{0.4cm} 
  \begin{subfigure}[t]{0.33\linewidth}
    \vspace{3.5mm}
    \centering
    \includegraphics[width=0.9\linewidth]{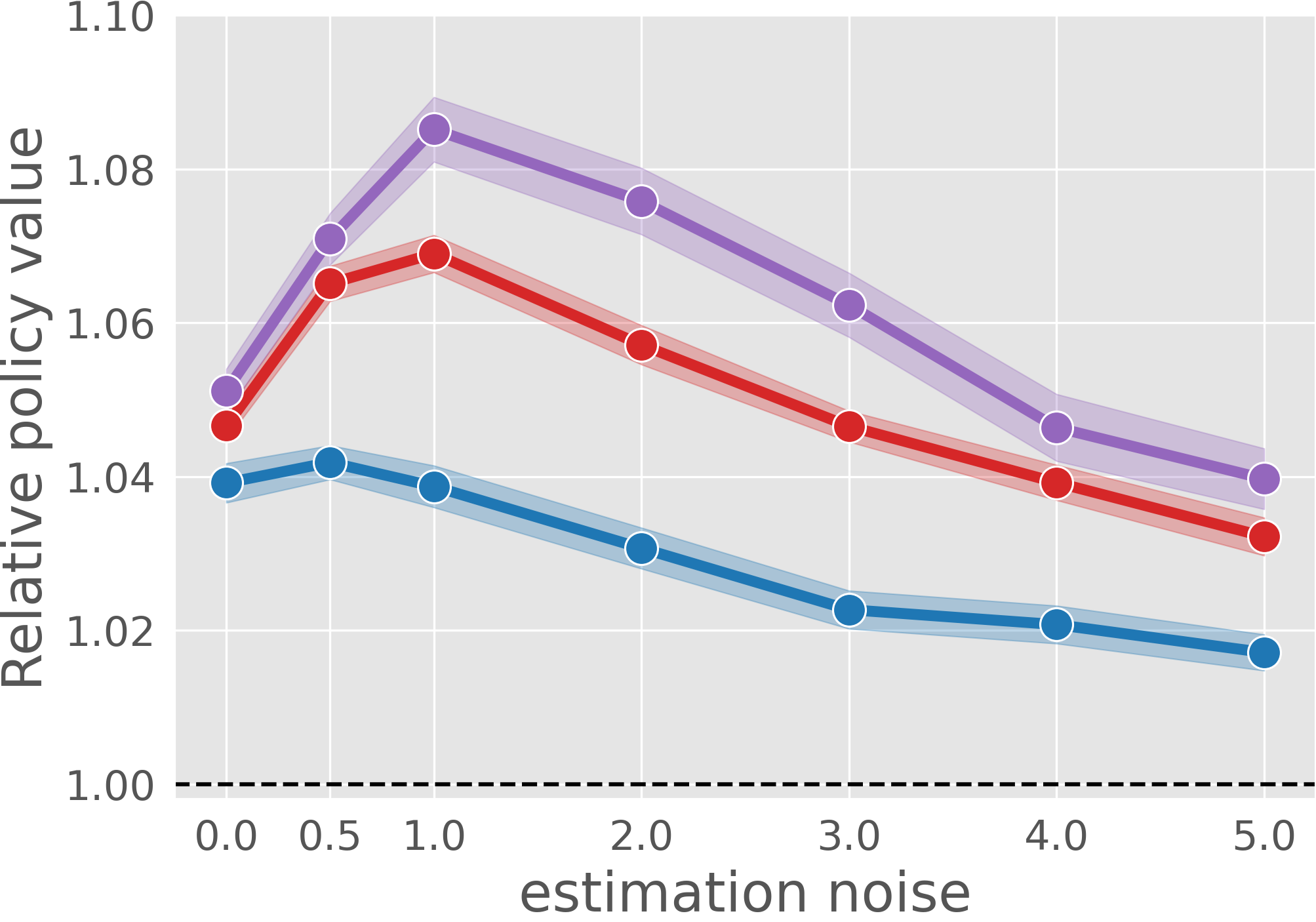}
    \vspace{-2mm}
    \caption{Relative policy values varying estimation noise}
    \label{fig:estimation_noise_KuaiRec}
    \vspace{-3mm}
  \end{subfigure}
  \caption{Comparisons of relative policy values with varying (a)the number of users and (b)estimation noise. The period $T$ is sufficiently large, and all items are sold out.}
\end{figure*}

\paragraph{Compared methods}
We compare OPLS with the conventional greedy method in Eq.~\eqref{eq:reg_supply}. In synthetic experiments except for Figure~\ref{fig:estimation_noise}, we use the true expected reward for both methods for the following reasons. First, we demonstrate OPLS can improve the policy value compared to the conventional greedy method that becomes the optimal policy with the true expected reward in the no limited supply setting. Second, this is the first research for OPL with limited supply, so we identify the situation where supply constraints have substantial effects on the performance of each method. 
Note that we also evaluate OPLS's performance with an estimated expected reward in the real-world experiments.

\subsection{Results}

We first demonstrate how OPLS chooses actions compared to the conventional greedy method using small-scale experiments. Then, we evaluate OPLS's performance in various settings. We compute 100 simulations with different random seeds to produce synthetic data instances. Unless otherwise specified, the number of actions is set to $|\calA|=100$, the popularity of actions is $\lambda = 0.5$ and $s_{\text{max}}=20$.
Except for Figure~\ref{fig:s_max}, the period $T$ is sufficiently large, and all items are sold out.

\paragraph{\textbf{How does OPLS choose actions compared to the conventional greedy method?}} 
To demonstrate the behavior of OPLS, we conduct a small-scale experiment where the number of users is set to $3$, the number of actions is $|\calA|=5$, the initial supply is random and $\lambda=0.0$. Table~\ref{table:small-scale_q(x,a)} reports the expected reward in this experiment. The actions with larger indices have higher expected rewards because of $\lambda=0.0$. Figure~\ref{fig:small-scall_timestep} shows the relative policy values between OPLS and the conventional greedy method $V(\pi_{\text{OPLS}})/V(\pi_{\text{greedy}})$ at each step. Note that the periods where the values remain constant indicate that the supply of all actions is zero. From Figure~\ref{fig:small-scall_timestep}, we can see that OPLS improves the final policy value compared to the conventional greedy method. 
Figure~\ref{fig:behavior_argmax} and \ref{fig:behavior_OPLS} illustrate how each method selects an action for a user. The value in each grid cell represents the proportion of the supply allocated to a specific user for a particular action at that time. From Figure~\ref{fig:behavior_argmax}, we observe that the conventional greedy method recommends actions with the highest expected rewards in the early time steps. This is why the conventional greedy method outperforms OPLS at the initial time step in Figure~\ref{fig:small-scall_timestep}. Eventually, the conventional greedy method recommends all actions evenly across all users. On the other hand, OPLS selectively recommends a specific action to a specific user. For example, approximately 80\% of $a_5$ is recommended to $x_1$. From Table~\ref{table:small-scale_q(x,a)}, $x_1$ has the highest expected reward for $a_5$ among users, so OPLS achieves effective recommendations in limited supply settings. It is also the case for the other user-action pairs. Thus, OPLS achieves a high policy value by selectively recommending actions to users with higher expected rewards.

\paragraph{\textbf{How does OPLS perform varying the action popularity?}}
Figure~\ref{fig:lambda} reports the relative policy values varying the action popularity ($\lambda$). The small values of $\lambda$ mean that all users likely prefer the actions in the same order. The results demonstrate that OPLS outperforms the conventional greedy method across all values of $\lambda$. Specifically, OPLS significantly improves policy values when $\lambda$ is small. This is because the more aligned the preferences are, the more crucial it is to recommend actions to the optimal users. In this situation, all users prefer the same action, so the conventional greedy method recommends actions evenly to all users. In contrast, OPLS selectively recommends the popular action to users with a high expected reward. Therefore, OPLS demonstrates a clear advantage in scenarios where many users favor a specific action.

Moreover, compared across the initial supply, OPLS improves policy values in the inverse proportional setting. This suggests that OPLS achieves superior performance when actions with high expected rewards are rare. This is because, in such situations, it is necessary to recommend the few valuable actions effectively.

\paragraph{\textbf{How does OPLS perform varying the the number of users?}}
Figure~\ref{fig:user_action_ratio} varies the number of users, while the number of actions is set to $|\calA| = 100$.
We observe that the relative policy values gradually increase as the number of users increases. 
This is because, as the number of users increases, items with the high expected reward become more scarce, making it more necessary to consider limited supply.

\paragraph{\textbf{How does OPLS perform varying estimation noises?}}
In Figure~\ref{fig:estimation_noise}, we use an estimated expected reward $\hat{q}(x,a) = q(x,a) + \mathcal{N}(0,\sigma)$ for both OPLS and the conventional greedy method. Figure~\ref{fig:estimation_noise} reports the relative policy value varying estimation noises $\sigma$. We observe that OPLS outperforms the conventional greedy method across various estimation noises, although its improvement gradually decreases as the estimation noise becomes larger. This result suggests that OPLS provides substantial improvements with a reasonably accurate estimate of the expected reward. 

\paragraph{\textbf{How does OPLS perform varying the max supply $s_{max}$?}}
Figure~\ref{fig:s_max} varies the max supply $s_{max}$ from 5 to 30. As the max supply $s_{max}$ increases, the limited supply setting gradually reduces to the \textbf{NO} limited supply setting. When the maximum supply is varied from 5 to 30, the proportion of unsold actions changes to approximately 0\%, 5\%, 30\%, 40\%, and 50\%.
We observe that OPLS improves the policy value across various max supplies. Specifically, OPLS provides substantial improvements when the max supply is small. This result is consistent with the results of Figure~\ref{fig:T=inf}. Moreover, OPLS exhibits performance equivalent to that of the conventional greedy method even when sufficient items are available. This suggests that OPLS provides substantial performance in both limited and no limited supply settings.

\section{Real-World Experiments}
This section demonstrates the effectiveness of OPLS in the limited supply setting on a subset of the real-world recommendation dataset called KuaiRec~\citep{gao2022kuairec}, which contains recommendation logs of the video-sharing app, Kuaishou. 
In our experiments, we use a subset of KuaiRec that contains 1,411 users and 3,327 items, for which user-item interactions are nearly fully observed (close to 100\% density). This property allows us to directly access the reward function. By leveraging this unique property, we can perform an OPL experiment on this dataset with a minimal synthetic component~\citep{gao2022kuairec}.
Since KuaiRec does not contain inventory information, we assign artificial inventory to each video, using the same method as in our synthetic data experiments. This allows us to evaluate OPLS's performance under limited supply. While this setup is purely a simulation for research purposes, introducing certain exposure constraints can also be relevant to real-world video recommendation systems. For example, constraints may be applied to limit the over-exposure of popular videos, ensure diversity across categories, or provide initial exposure guarantees for new items to address cold-start issues.

To perform an OPL experiment on this dataset, we sample 1,000 users and 1,000 items, and treat their interactions as the action value $q_r(x,a)$. We set the click probability $q_c(x,a)=1$ for every user–item pair. We then define the logging policy following Eq.~\eqref{eq:pi_0}, where we set $\beta = 1.0$. We obtain an estimated expected reward $\hat{q}(x,a)$ using a 3-layer neural network and the logged data $\calD$.

\paragraph{\textbf{Results}}
In figure~\ref{fig:n_users_estimate_KuaiRec}, we vary the number of users. We fixed the number of actions $|\calA| = 1000$. We observe that OPLS improves the policy value across various numbers of users. The results suggest that OPLS provides substantial improvements in more realistic situations where we have no access to the true expected reward.

Figure~\ref{fig:estimation_noise_KuaiRec}, we use estimated expected reward $\hat{q}(x,a) = q(x,a) + \mathcal{N}(0,\sigma)$ for both OPLS and the conventional greedy method. We observe that OPLS outperforms the conventional greedy method across various estimation noises. Unlike Figure~\ref{fig:estimation_noise}, the relative policy value increases when estimation noises are small. This shows that OPLS could be more robust against estimation noise than the conventional greedy method. 

\section{Conclusion and Future Work}
In this paper, we addressed the problem of off-policy learning (OPL) with limited supply. 
Existing OPL methods do not account for inventory constraints, which can lead to suboptimal allocations since selecting the item that maximizes the expected reward for one user may forgo the opportunity to obtain a higher expected reward in the future. To overcome this limitation, we proposed Off-Policy Learning with Limited Supply (OPLS), which selects items by maximizing the relative expected reward across users. Moreover, OPLS can incorporate supply prediction, enabling it to adaptively choose items according to supply conditions. Through experiments on both synthetic and real-world datasets, we demonstrated that OPLS consistently outperforms existing methods with limited supply.  

While this paper establishes the effectiveness of OPLS under limited supply, several directions remain open for future research. In this work, we addressed the case where each item has an individual supply constraint. A natural extension is to consider global constraints that apply to the entire set of items, which would capture more complex allocation scenarios. Moreover, since inventory is limited, future work may need to incorporate fairness-aware objectives into OPLS to balance efficiency with equity among users, rather than focusing solely on revenue maximization.

%% file: appendix.tex
\newpage
\appendix
\onecolumn
\raggedbottom
\allowdisplaybreaks

\section{Calculation of \emph{greedy} in Table~\ref{table:coupon_example}}
\label{app:average-greedy}

We present the detailed calculation of the greedy method in Table~\ref{table:coupon_example}.

Let $\mathcal{X}=\{x_1,x_2,x_3\}$ and $\mathcal{A}=\{\text{30\%OFF},\text{50\%OFF},\text{70\%OFF}\}$ with expected rewards $q(x,a)$ given in Table~\ref{table:coupon_example}.
Let $\sigma = (u_1, u_2, u_3)$ denote an arrival order of users, where $\sigma$ is one of the $3! = 6$ possible permutations of $\mathcal{X}$.
A \emph{greedy} allocation sequentially assigns to each arriving user $u_t$ the remaining coupon $a \in \mathcal{A}$ that maximizes $q(u_t,a)$.
The total reward under order $\sigma$ is then given by $\sum_{x \in \mathcal{X}} q(x,a_\sigma)$, where $a_\sigma$ denotes the coupon allocated to $x$ under $\sigma$.
Assuming all six arrival orders occur uniformly at random, we compute the policy value as
\begin{equation*}
    V(\pi) = \frac{1}{6}\sum_{\sigma} \sum_{x} q(x, a_\sigma).
\end{equation*}

We enumerate all six permutations $\sigma$ and record the greedy choices and resulting rewards:

\begin{center}
\begin{tabular}{c|l|c}
\toprule
Arrival order $\sigma$ & \hspace{0.8em}Greedy allocation $(a, q(x,a))$ & $\sum_x q(x,a_\sigma)$ \\
\midrule
$(x_1,x_2,x_3)$ & $x_1\!\to\!50\%\text{OFF}\,(250),\;\; x_2\!\to\!70\%\text{OFF}\,(120),\;\; x_3\!\to\!30\%\text{OFF}\,(60)$ & $430$ \\
$(x_1,x_3,x_2)$ & $x_1\!\to\!50\%\text{OFF}\,(250),\;\; x_3\!\to\!70\%\text{OFF}\,(70), \;\;\;\; x_2\!\to\!30\%\text{OFF}\,(100)$ & $420$ \\
$(x_2,x_1,x_3)$ & $x_2\!\to\!50\%\text{OFF}\,(280),\;\; x_1\!\to\!70\%\text{OFF}\,(200),\;\; x_3\!\to\!30\%\text{OFF}\,(60)$ & $540$ \\
$(x_2,x_3,x_1)$ & $x_2\!\to\!50\%\text{OFF}\,(280),\;\; x_3\!\to\!70\%\text{OFF}\,(70), \;\;\;\; x_1\!\to\!30\%\text{OFF}\,(80)$  & $430$ \\
$(x_3,x_1,x_2)$ & $x_3\!\to\!50\%\text{OFF}\,(100),\;\; x_1\!\to\!70\%\text{OFF}\,(200),\;\; x_2\!\to\!30\%\text{OFF}\,(100)$ & $400$ \\
$(x_3,x_2,x_1)$ & $x_3\!\to\!50\%\text{OFF}\,(100),\;\; x_2\!\to\!70\%\text{OFF}\,(120),\;\; x_1\!\to\!30\%\text{OFF}\,(80)$  & $300$ \\
\bottomrule
\end{tabular}
\end{center}

Therefore, we calculate the policy value of the greedy method as
\[
V(\pi) = \frac{1}{6}\sum_{\sigma}\sum_{x} q\bigl(x,a_\sigma\bigr)
\;=\;
\frac{430+420+540+430+400+300}{6}
= 420.
\]

\section{omitted proof}
Here, we provide the derivations and proofs that are omitted in the main text.
\subsection{Derivation of policy value with limited supply $V(\pi)$ in Eq~\eqref{eq:V_pi_modified}}
\label{derivation of policy value}
\begin{align}
    V(\pi) &= \mathbb{E}_{p(s_1) \prod_{t=1}^T p(x_t)\pi(a_t|x_t,s_t)p(c_t|x_t,a_t)p(r_t|x_t,a_t)p(s_{t+1}|s_t, a_t, c_t)}\left[ \sum_{t=1}^T c_t r_t \right] \notag \\
    &= \mathbb{E}_{p(s_1)}\Big[ \sum_{x_1}p(x_1)\sum_{a_1}\pi(a_1|x_1,s_1)\sum_{c_1}p(c_1|x_1,a_1)\sum_{r_1}p(r_1|x_1,a_1)\sum_{s_2}p(s_2|s_1,a_1, c_1)c_1r_1 \notag \\
    &\quad\quad + \sum_{x_2}p(x_2)\sum_{a_2}\pi(a_2|x_2,s_2)\sum_{c_2}p(c_2|x_2,a_2)\sum_{r_2}p(r_2|x_2,a_2)\sum_{s_3}p(s_3|s_2, a_2, c_2)c_2r_2 + \cdots \Big]\notag \\
    &= \mathbb{E}_{p(s_1)}\Big[ \sum_{x_1}p(x_1)\sum_{a_1}\pi(a_1|x_1,s_1)
    p(c_1=1|x_1,a_1)\sum_{r_1}p(r_1|x_1,a_1)\sum_{s_2}p(s_2|s_1,a_1, c_1=1)r_1 \notag \\
    &\quad\quad + \sum_{x_2}p(x_2)\sum_{a_2}\pi(a_2|x_2,s_2)p(c_2=1|x_2,a_2)\sum_{r_2}p(r_2|x_2,a_2)\sum_{s_3}p(s_3|s_2, a_2, c_2=1)r_2 + \cdots \Big] \label{c_binary}\\
    &= \mathbb{E}_{p(s_1)}\Big[ \sum_{x_1}p(x_1)\sum_{a_1}\pi(a_1|x_1,s_1)
    p(c_1=1|x_1,a_1)\sum_{r_1}p(r_1|x_1,a_1)r_1 \notag \\
    &\quad\quad + \sum_{x_2}p(x_2)\sum_{a_2}\pi(a_2|x_2,s_2)p(c_2=1|x_2,a_2)\sum_{r_2}p(r_2|x_2,a_2)r_2 + \cdots \Big] \label{p(s_t+1)_deterministic}\\
    &= \mathbb{E}_{p(s_1)}\Big[ \sum_{x_1}p(x_1)\sum_{a_1}\pi(a_1|x_1,s_1)
    p(c_1=1|x_1,a_1)q_{r_1}(x_1,a_1) \notag \\
    &\quad\quad + \sum_{x_2}p(x_2)\sum_{a_2}\pi(a_2|x_2,s_2)p(c_2=1|x_2,a_2)q_{r_2}(x_2,a_2) + \cdots \Big]\notag \\
    &= \mathbb{E}_{p(s_1)}\Big[ \sum_{t=1}^T\sum_{x_t}p(x_t)\sum_{a_t}\pi(a_t|x_t,s_t)
    p(c_t=1|x_t,a_t)q_{r_t}(x_t,a_t) \Big]\notag \\
    &= \sum_{t=1}^T \mathbb{E}_{p(s_1)p(x_t)\pi(a_t|x_t,s_t)}\Big[ 
    q_{c_t}(x_t,a_t)q_{r_t}(x_t,a_t) \Big]\notag \\
    &= \sum_{t=1}^T \mathbb{E}_{p(s_1)p(x_t)\pi(a_t|x_t,s_t)}\Big[ 
    q(x_t,a_t) \Big]\notag \\
    &= \mathbb{E}_{p(s_1)}\Big[ \sum_{t=1}^T\sum_{x_t}p(x_t)\sum_{a_t}\pi(a_t|x_t,s_t)
    p(c_t=1|x_t,a_t)q_{r_t}(x_t,a_t) \Big]\notag \\
    &= \sum_{t=1}^T \mathbb{E}_{p(s_1)p(x_t)\pi(a_t|x_t,s_t)}\left[ 
    q(x_t,a_t) \right],\notag 
\end{align}
where we utilized that $c$ is binary, and $p(s_{t+1}|s_t, a_t, c_t=1)$ is deterministic in Eq.~\eqref{c_binary} and \eqref{p(s_t+1)_deterministic}, respectively.

\subsection{Derivation of $V(\pi_\text{greedy})$ in Eq.~\eqref{eq:V_pi_greedy}}
\label{derivation of greedy}
\begin{align}
    V(\pi_\text{greedy}) 
    &= \sum_{t=1}^T \mathbb{E}_{p(s_1)p(x)\pi_\text{greedy}(a|x,s_t)}\left[ q(x,a) \right] 
    = \sum_{t=1}^T \mathbb{E}_{p(s_1)p(x)}\left[ \sum_{a \in \calA_{s_t}} \pi_\text{greedy}(a|x,s_t) q(x,a) \right]\notag \\
    &= \sum_{t=1}^T \mathbb{E}_{p(s_1)p(x)}\left[ \sum_{a \in \calA_{s_t}} q(x,a) \indicator{a=a^{(t)}} \right]
    = \sum_{t=1}^T \mathbb{E}_{p(s_1)p(x)}\left[ q(x,a^{(t)})  \right]\notag\\
    &= \sum_{k=1}^K \mathbb{E}_{p(x)}\left[ q(x,a_k)  \right], \label{eq:derivation_V_greedy}
\end{align}
where we use $p(s_1)=1$ and $a^{(t)}=a_k$ and in Eq.~\eqref{eq:derivation_V_greedy}.

\subsection{Derivation of $V(\pi_{j,k})$ in Eq.~\eqref{eq:V_pi_jk}}
\label{derivation of pi_jk}
\begin{align*}
    V(\pi_{j,k}) 
    &= p(x_j) \left\{ q(x_j, a_k) + V(\pi_\text{greedy},\calA/a_k) \right\} 
    + \sum_{x \in \calX / x_j} p(x)\left\{ q(x, a_1) + V(\pi_\text{greedy},\calA/a_1) \right\} \\
    &= p(x_j) \left\{ q(x_j, a_k) + V(\pi_\text{greedy}) - \mathbb{E}_{p(x)}[q(x,a_k)] \right\} 
    + \sum_{x \in \calX / x_j} p(x)\left\{ q(x, a_1) + V(\pi_\text{greedy}) - \mathbb{E}_{p(x)}[q(x,a_1)] \right\} \\
    &= p(x_j) \left\{ q(x_j, a_k) + V(\pi_\text{greedy}) - \mathbb{E}_{p(x)}[q(x,a_k)] \right\} \\
    &\quad\quad\quad + \cancel{\mathbb{E}_{p(x)} \left[ q(x, a_1)\right]} + V(\pi_\text{greedy}) - \cancel{\mathbb{E}_{p(x)}[q(x,a_1)]}  \\
    &\quad\quad\quad\quad\quad\quad - p(x_j)\left\{ q(x_j, a_1) + V(\pi_\text{greedy}) - \mathbb{E}_{p(x)}[q(x,a_1)] \right\}\\
    &= p(x_j) \left\{ q(x_j,a_k) - \mathbb{E}_{p(x)}[q(x,a_k)] - (q(x_j,a_1) - \mathbb{E}_{p(x)}[q(x,a_1)]) \right\} + V(\pi_\text{greedy}),
\end{align*}
where $V(\pi_{\text{greedy}},\calA/a_k') = \sum_{k=1}^K \mathbb{E}_{p(x)}[q(x, a_k)]  \indicator{k\neq k'}$.

\section{EXTENTION TO FAIRNESS}
Future work may need to incorporate fairness-aware
objectives into OPLS to balance efficiency with equity among users, rather than focusing solely on revenue maximization. While addressing such fairness concerns is outside the scope of this paper, if one aims to deal with it, it is indeed possible to extend OPLS to ensure fairness across users by introducing a weighting parameter, $\beta \in [0,1]$, into the decision rule. Specifically, we can modify OPLS to select the action that maximizes the estimated reward minus a beta-weighted term representing the average reward across all users: 
\begin{align*}
    \pi_{\text{OPLS}}(a_t | x_t, s_t) =  
    \left\{
    \begin{aligned}
        1  \quad & \text{if } a_t = \underset{a \in \mathcal{A}_{s_t}}{\operatorname{argmax}} \left\{ \hat{q}(x_t, a) - \beta \cdot \frac{1}{n} \sum_{i=1}^n \hat{q}(x_i, a) \right\} \\
        0 \quad & \quad \text{otherwise}
    \end{aligned}
    \right. .
\end{align*}
This allows beta to smoothly control the trade-off between individual optimality and global optimality.

\section{DETAILED EXPERIMENT SETTINGS AND RESULTS}
\label{additional_result}
\subsection{Synthetic Experiments}

\textbf{Detailed Setup.} \quad 
We describe synthetic experiment settings in detail. In the synthetic experiments, we define a click probability $q_c(x,a)$ and an action value $q_r(x,a)$ as follows.
\begin{align*}
    q_c(x,a) = \lambda \cdot f_c(x,a) + (1-\lambda) \cdot g_c(x,a) ,\\
    q_r(x,a) = \lambda \cdot f_r(x,a) + (1-\lambda) \cdot g_r(x,a), 
\end{align*}
where $g(x,a)$ is a reward function that satisfies $g(x,a_k) \ge g(x,a_{k+1})$ for all users. 
We synthesize $f_c$ and $f_r$ using \textbf{obp.dataset.logistic\_reward\_function} and \textbf{obp.dataset.linear\_reward\_function} from OpenBanditPipeline~\citep{saito2021open}, respectively. To construct $g$, we then sample elements of $g$ from a uniform distribution with range $[0, \max{(f)}]$, and sort them in descending order.
\\ \\
\begin{wrapfigure}{r}{0.38\textwidth}
    \vspace{-5mm}
    \centering
    \includegraphics[scale=0.21]{image/legend.png} \\
    \vspace{1mm}
    \Description{}
    \centering
    \includegraphics[width=0.82\linewidth]{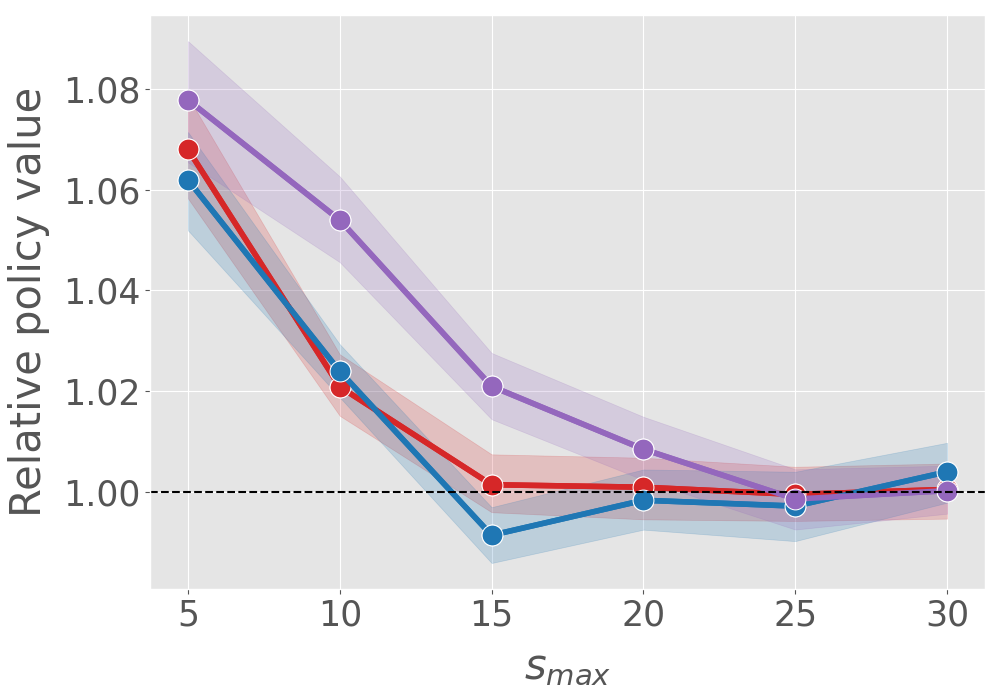}
    \vspace{-3mm}
    \caption{Relative policy values varying max supply ($s_{max}$)}
    \label{fig:s_max_naive}
    \vspace{-5mm}
\end{wrapfigure}
\textbf{Additional Result.} \quad Figure~\ref{fig:s_max_naive} reports relative policy values varying the maximum supply $s_{max}$ with naive estimation of $\mathcal{A}_{\text{sold}}$ and $\mathcal{A}_{\text{unsold}}$. The number of actions is set to $|\calA|=100$, the popularity of actions is $\lambda = 0.5$ and $T=2500$.

In Figure~\ref{fig:s_max_naive}, we observe that the relative policy values gradually decrease and converge to 1. This result is consistent with Figure~\ref{fig:s_max}, even though OPLS' performance with the naive estimation is slightly worse than with the straightforward estimation. Figure~\ref{fig:s_max_naive} suggests that OPLS with the naive estimation of sold actions provides substantial improvements in policy values with a reduction of computational costs.

\subsection{Real-World Experiments}
\textbf{Detailed Setup.} \quad
We describe the real-world experiment settings on KuaiRec~\citep{gao2022kuairec} in detail. KuaiRec has user features and
user-item interactions which are almost fully observed with nearly 100\% density for the subset of its users and items. We consider user features as contexts $x$, where we reduce feature  dimensions based on PCA implemented in scikit-learn~\citep{pedregosa2011scikit}. Then, we consider the user-item interactions $r$ as the action value $q_r(x,a)$. Since users swipe past videos without clicking in the video-sharing app, we set the click probability $q_c(x,a)$ to $1$ for every user–item pair $(x,a)$. In this setting, we sample a reward $r$ from a truncated normal distribution with mean $q_r(x,a)$ and standard deviation $\sigma = 1.0$.

\begin{figure*}[h]
    \centering
    \vspace{3mm}
    \includegraphics[scale=0.3]{image/legend.png} \\
    \vspace{-5mm}
    \Description{}
  \begin{subfigure}[t]{0.35\linewidth}
    \vspace{3mm}
    \centering
    \includegraphics[width=0.9\linewidth]{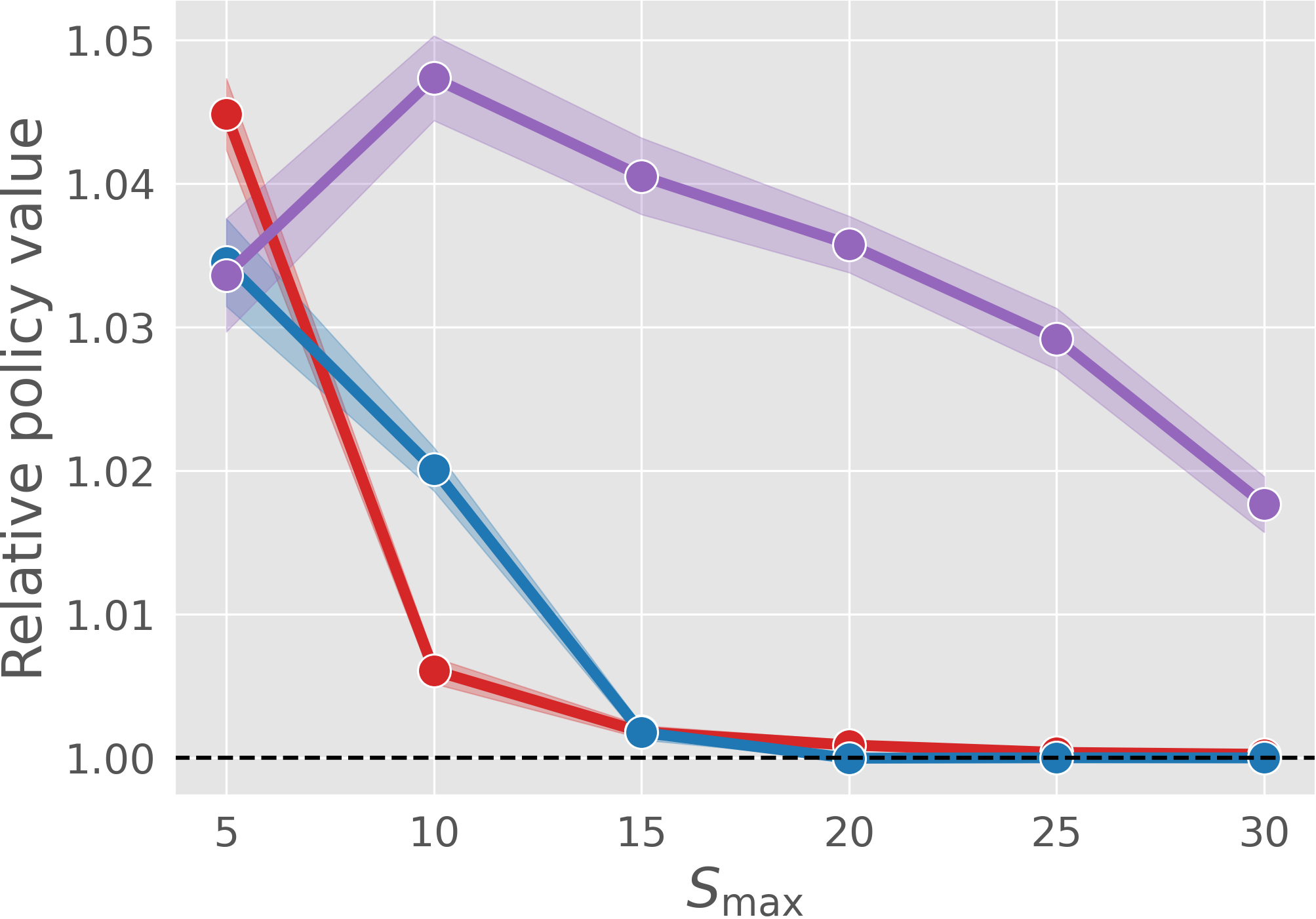}
    \vspace{-3mm}
    \caption{Relative policy values varying max supply ($s_{max}$)}
    \label{fig:s_max_true_KuaiRec}
  \end{subfigure}
  \hspace{0.4cm} 
  \begin{subfigure}[t]{0.36\linewidth}
    \vspace{1.5mm}
    \centering
    \includegraphics[width=0.9\linewidth]{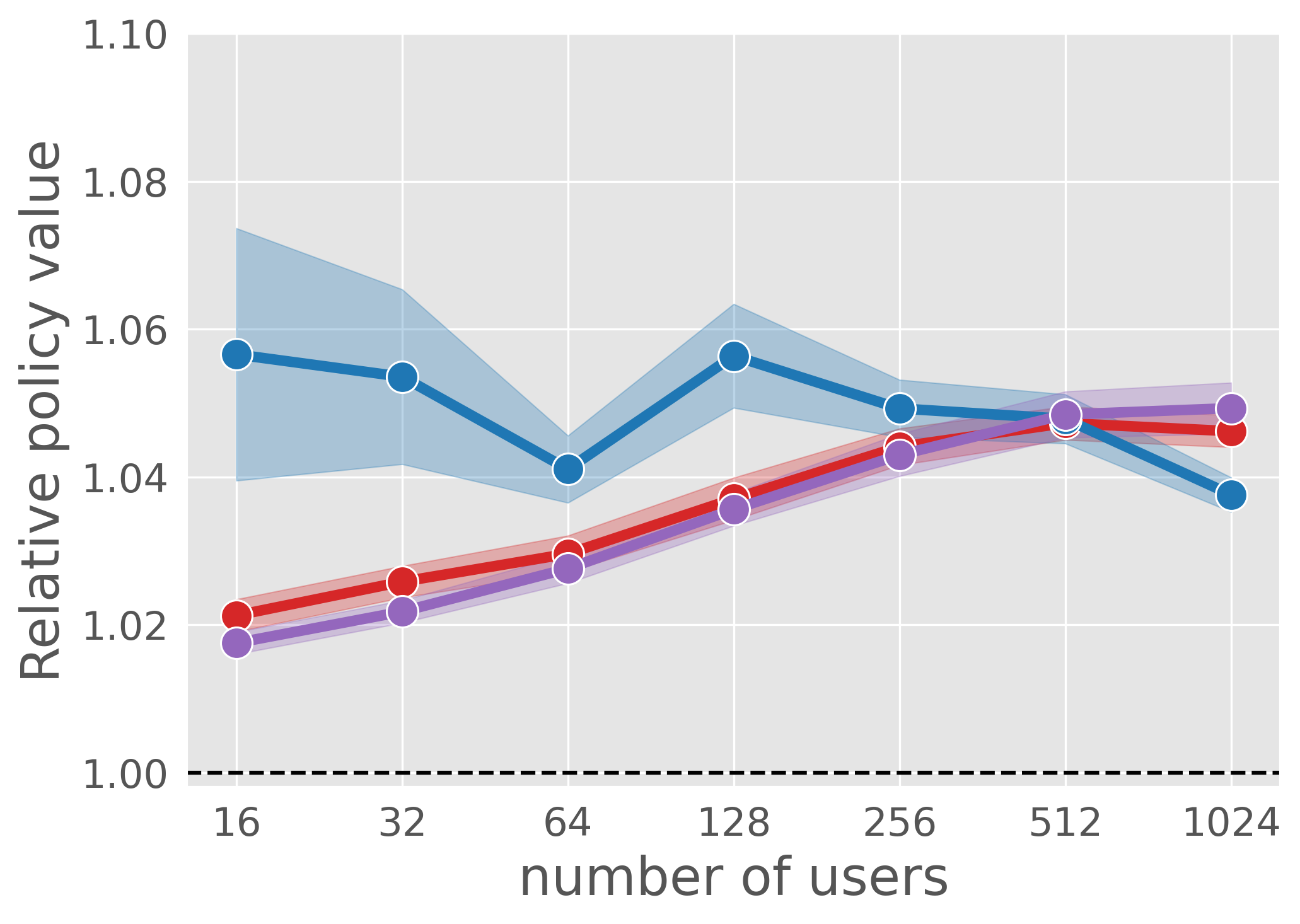}
    \vspace{-3mm}
    \caption{Relative policy values varying the number of users}
    \label{fig:n_users_true_KuaiRec}
  \end{subfigure}
  \caption{Comparisons of relative policy values with varying (a) max supply ($s_{max}$) and (b) the number of users using the true expected reward $q(x,a)$ in both cases. For (b), the period $T$ is sufficiently large and all items are sold out.}
\end{figure*}

We define the logging policy based on the expected reward function $q(x,a)$ as follows.
\begin{align}
    \pi_0(a_t|x_t,s_t) = \frac{\exp{(\beta \cdot \hat{q}(x_t,a_t))}}{\sum_{a' \in \calA_{s_t}} \exp{(\beta \cdot \hat{q}(x_t,a'))}},
    \label{eq:pi_0_app}
\end{align}
where we use $\beta = 1.0$ , $\hat{q}(x,a) = q(x,a) + \mathcal{N}(0, 5.0)$ and $q(x,a) = q_c(x,a) \cdot q_r(x,a)$.
\\ \\
\textbf{Additional results.} \quad
In Figure~\ref{fig:s_max_true_KuaiRec}, we  vary the max supply $s_{max}$ from 5 to 30. For computational efficiency, we employ the naive method in this experiment. When the max supply $s_{\text{max}}$ is small (e.g., 5 or 10), almost all items are sold out. As $s_{\text{max}}$ increases further, the supply constraint becomes less restrictive, and the limited-supply setting gradually reduces to the \textbf{NO} limited supply setting. We observed that even when the naive method is employed, the improvement becomes more significant when the max supply $s_{max}$ is small. The naive method  demonstrates performance equivalent to that of the conventional greedy method, when sufficient items are available. 


Figure~\ref{fig:n_users_true_KuaiRec}, we  vary the number of users. We fixed the number of actions $|\calA| = 1000$. We observe a trend similar to that in figure~\ref{fig:user_action_ratio}, where the relative policy values gradually increase as the number of users increases. This is because, as the number of users increases, high-reward items that improve the policy value become scarce, emphasizing the need to consider limited supply.

